\pdfoutput=1
\documentclass[10pt,twocolumn,letterpaper]{article}

\usepackage[pagenumbers]{cvpr}   %

\usepackage{cuted}

\usepackage{currfile}

\usepackage{caption}

\usepackage{amsthm}
\usepackage{bm}
\usepackage{pifont}
\usepackage{multirow}
\usepackage{makecell}
\usepackage{array}
\newtheorem{proposition}{Proposition}

\newcommand{\methodname}{Commit2Drive}
\newcommand{\method}{\methodname\xspace}

\newcommand{\papertitle}{\methodname: Perception-Grounded Language Decisions as Commitments\\for Trajectory Planning in Vision-Language-Action Driving Models}

\newcommand{\codebook}{Maneuver Bank\xspace}

\newcommand{\SuppFormalRef}{D.3}    %
\newcommand{\SuppDataRef}{C}        %
\newcommand{\SuppAnalysisRef}{E.2}  %
\newcommand{\SuppParamsRef}{D}      %
\newcommand{\SuppAblationRef}{E.3}  %

\newcommand{\SuppData}{Sec.~\SuppDataRef{} of the supplementary material\xspace}
\newcommand{\SuppAnalysis}{Sec.~\SuppAnalysisRef{} of the supplementary material\xspace}

\newcommand{\SuppAblation}{Sec.~\SuppAblationRef{} of the supplementary material\xspace}

\newcommand{\TBD}[1]{\textcolor{red}{\if\relax\detokenize{#1}\relax\texttt{--}\else#1\fi}}
\newcommand{\prov}[1]{\textcolor{Plum}{#1}}
\newcommand{\repro}[1]{\textit{#1}}

\newcommand{\dif}[1]{\,{\scriptsize\bfseries(#1)}}

\definecolor{draftblue}{RGB}{35,83,126}
\definecolor{draftlight}{RGB}{243,247,250}

\DeclareMathOperator*{\argmax}{arg\,max}

\newcommand{\Dset}{\mathcal{D}}

\newcommand{\Dadm}{\mathcal{D}_{F}}

\usepackage{microtype}
\renewcommand{\paragraph}[1]{\vspace{.5em}\noindent\textbf{#1}}
\makeatletter
\@ifundefined{algorithmic}{}{\makeatother }

\usepackage{float}
\floatstyle{ruled}
\newfloat{algorithm}{tbp}{loa}
\floatname{algorithm}{Algorithm}

\newcounter{ALG@line}
\newcount\ALG@depth
\newdimen\ALG@indent \ALG@indent=1.2em   %
\newdimen\ALG@numwd  \ALG@numwd=1.7em    %
\newif\ifALG@numbered

\newenvironment{algorithmic}[1][0]{%
  \par
  \ifnum#1>\z@ \ALG@numberedtrue \else \ALG@numberedfalse \fi
  \setcounter{ALG@line}{0}\global\ALG@depth=\z@
  \footnotesize\raggedright
  \setlength{\parindent}{\z@}\setlength{\parskip}{\z@}%
}{\par}

\newcommand{\ALG@startline}{%
  \par
  \hangindent=\dimexpr\ALG@numwd+\ALG@depth\ALG@indent+\ALG@indent\relax
  \hangafter=\@ne
  \noindent
  \ifALG@numbered
    \stepcounter{ALG@line}%
    \makebox[\ALG@numwd][r]{\scriptsize\theALG@line:\hspace{0.4em}}%
  \else
    \hspace*{\ALG@numwd}%
  \fi
  \hspace*{\ALG@depth\ALG@indent}%
  \ignorespaces}
\newcommand{\ALG@open}{\global\advance\ALG@depth\@ne}
\newcommand{\ALG@close}{\ifnum\ALG@depth>\z@\global\advance\ALG@depth\m@ne\fi}

\newcommand{\algorithmicrequire}{\textbf{Input:}}
\newcommand{\algorithmicensure}{\textbf{Output:}}

\newcommand{\State}{\ALG@startline}
\newcommand{\Statex}{\par\noindent\hspace*{\ALG@numwd}\hspace*{\ALG@depth\ALG@indent}\ignorespaces}
\newcommand{\Require}{\ALG@startline\algorithmicrequire\ \ignorespaces}
\newcommand{\Ensure}{\ALG@startline\algorithmicensure\ \ignorespaces}
\newcommand{\Trainable}{\ALG@startline\textbf{Trainable:}\ \ignorespaces}   %
\newcommand{\For}[1]{\ALG@startline\textbf{for} #1 \textbf{do}\ALG@open}
\newcommand{\ForAll}[1]{\ALG@startline\textbf{for all} #1 \textbf{do}\ALG@open}
\newcommand{\EndFor}{\ALG@close\ALG@startline\textbf{end for}}
\newcommand{\While}[1]{\ALG@startline\textbf{while} #1 \textbf{do}\ALG@open}
\newcommand{\EndWhile}{\ALG@close\ALG@startline\textbf{end while}}
\newcommand{\If}[1]{\ALG@startline\textbf{if} #1 \textbf{then}\ALG@open}
\newcommand{\ElsIf}[1]{\ALG@close\ALG@startline\textbf{else if} #1 \textbf{then}\ALG@open}
\newcommand{\Else}{\ALG@close\ALG@startline\textbf{else}\ALG@open}
\newcommand{\EndIf}{\ALG@close\ALG@startline\textbf{end if}}

\newcommand{\Phase}[1]{\ALG@startline{\color{black!55}\#~#1}}
\makeatother

\usepackage{tcolorbox}
\definecolor{promptblue}{RGB}{0,0,205}
\newtcolorbox{promptbox}{colback=black!5, colframe=black!65, boxrule=0.9pt, arc=2.5mm,
  left=5mm, right=5mm, top=2mm, bottom=2mm, fontupper=\footnotesize}
\newcommand{\prompthead}[1]{{\bfseries\color{promptblue}#1}\par}  %
\newcommand{\promptsep}{\par\vspace{4pt}\hrule height 0.4pt\vspace{4pt}} %
\newcommand{\slot}[1]{\textlangle\textit{#1}\textrangle}             %

\newcounter{suppquestion}
\newcommand{\question}[1]{\par\vspace{.6em}\refstepcounter{suppquestion}%
  \noindent\textbf{Q\thesuppquestion.}\ \textit{#1}\par\nopagebreak}

\newcommand{\suppnumbering}{%
  \renewcommand{\thesection}{\Alph{section}}%
  \renewcommand{\thetable}{A\arabic{table}}%
  \renewcommand{\thefigure}{A\arabic{figure}}%
  \renewcommand{\theequation}{A\arabic{equation}}%
  \renewcommand{\thealgorithm}{A\arabic{algorithm}}%
  \renewcommand{\theproposition}{A\arabic{proposition}}}

\makeatletter
\newcommand{\suppanchor}[2]{%
  \begingroup
  \edef\supp@want{#1}\edef\supp@have{#2}%
  \ifx\supp@want\supp@have\else
    \PackageWarningNoLine{supp}{The main paper cites `Sec. \supp@want' of the supplementary
      material, but that content is now Sec. \supp@have. Update the \string\Supp...Ref
      macros in preamble.tex}%
    \textcolor{red}{\bfseries\,[main paper cites Sec.~\supp@want, this is Sec.~\supp@have]}%
  \fi
  \endgroup}
\makeatother

\ifdefined\pdfsuppressptexinfo \pdfsuppressptexinfo=-1 \fi
\ifdefined\pdfinfoomitdate \pdfinfoomitdate=1 \fi
\ifdefined\pdftrailerid \pdftrailerid{} \fi

\AtEndPreamble{%
  \crefname{algorithm}{Alg.}{Algs.}\Crefname{algorithm}{Algorithm}{Algorithms}%
  \crefname{proposition}{Prop.}{Props.}\Crefname{proposition}{Proposition}{Propositions}%
  \crefformat{suppquestion}{Q#2#1#3}\Crefformat{suppquestion}{Q#2#1#3}}

\renewcommand{\prov}[1]{#1}

\renewcommand{\SuppFormalRef}{\ref{supp:sec:shortcut}}
\renewcommand{\SuppDataRef}{\ref{supp:sec:data}}
\renewcommand{\SuppAnalysisRef}{\ref{supp:sec:gapfull}}
\renewcommand{\SuppParamsRef}{\ref{supp:sec:impl}}
\renewcommand{\SuppAblationRef}{\ref{supp:sec:ablation}}
\renewcommand{\suppanchor}[2]{}

\definecolor{cvprblue}{rgb}{0.21,0.49,0.74}
\usepackage[pagebackref,breaklinks,colorlinks,allcolors=cvprblue]{hyperref}
\makeatletter
\@ifpackageloaded{cleveref}{}{\usepackage[capitalize]{cleveref}}
\makeatother

\def\confName{CVPR}
\def\confYear{2027}

\title{\papertitle}

\author{Zikai Zhang\\
{\normalsize Department of Computer Science}\\
{\normalsize Durham University}\\
{\tt\small zikai.zhang@durham.ac.uk}
\and
Hubert P. H. Shum\\
{\normalsize Department of Computer Science}\\
{\normalsize Durham University}\\
{\tt\small hubert.shum@durham.ac.uk}
\and
Toby Breckon\textsuperscript{\dag}\\
{\normalsize Department of Computer Science}\\
{\normalsize Durham University}\\
{\tt\small toby.breckon@durham.ac.uk}
}

\begin{document}
\maketitle
{\let\thefootnote\relax\footnotetext{\textsuperscript{\dag}Corresponding author.}}
\begin{strip}
\centering
\includegraphics[width=\linewidth]{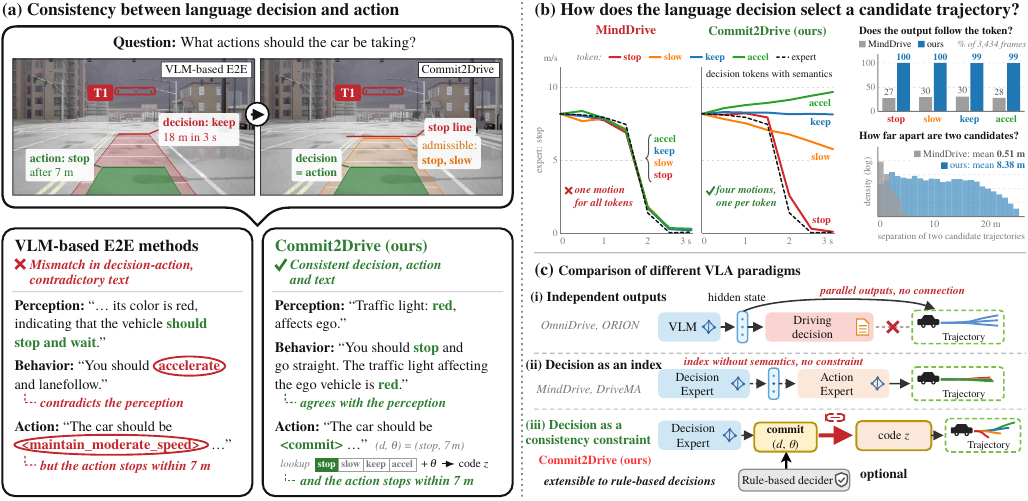}
\captionsetup{hypcap=false}
\captionof{figure}{\textbf{Comparison of \method with existing VLA driving models.} (a) Existing VLM-based models state a decision that contradicts their own action, while the decision and the action of \method are consistent. (b) The decision tokens of MindDrive lead to nearly the same trajectory, while those of \method lead to distinct motions. (c) Paradigms of VLA models: independent outputs, the decision as an index, and the decision as a consistency constraint (ours).}
\captionsetup{hypcap=true}
\label{fig:\currfilebase}%
\end{strip}

\begin{abstract}
Vision-Language-Action (VLA) models introduce language reasoning into end-to-end autonomous driving. They output a language decision together with a trajectory, which makes the planner more interpretable.
However, in existing methods the language decision is either an independent output or an index that selects among similar candidate trajectories, and its annotations are not verified against the trajectory.
As a result, the language decision and the trajectory are often inconsistent in closed-loop driving.
To address this, we propose \method, a framework that imposes the language decision as a consistency constraint on trajectory planning.
The decision is a commitment, a maneuver class with a continuous parameter that one deterministic rule reads from any trajectory, and it is grounded in facts reported by the perception of the model itself.
A \codebook maps the commitment to a code with a kinematic meaning, and consistency losses train the planner to execute every admissible decision of a scene, so that the trajectory follows the decision.
An automatic data pipeline supplies the supervision. Decisions are labeled from the future motion of the expert, counterfactual executions train the decision and the trajectory together, and rule-based functions with a Vision-Language Model (VLM) generate, review and refine the language annotations.
The code is a lookup that involves no hidden state of the language model, so \method is extensible to rule-based decisions, which the same planner executes without retraining.
\method uses a 0.5B language model and runs closed-loop inference online on a single consumer graphics card.
On Bench2Drive, \method reaches a Driving Score (DS) of 78.50 and a Success Rate (SR) of 55.91\% with a lightweight model, which is competitive with the state of the art, and greatly reduces the language-action inconsistency.
Code will be released soon.
\end{abstract}

\section{Introduction}
\label{sec:intro}

Autonomous driving requires a model to understand complex traffic scenes and to plan safe trajectories.
Traditional End-to-End (E2E) methods~\cite{transfuser,uniad,vad,drivetransformer} integrate perception, prediction and planning into one differentiable network and map sensor inputs directly to trajectories.
These methods have made great progress, but they work as black boxes and give no explanation for the trajectory they output.
With the rapid progress of Vision-Language Models (VLMs) in visual understanding and reasoning, recent work extends them to Vision-Language-Action (VLA) models for driving~\cite{drivevlm,emma,orion,simlingo,autovla,minddrive}.
A VLA model outputs chain-of-thought reasoning or a language decision together with the trajectory.
This paradigm improves generalization with the world knowledge of the VLM, and it lets human drivers understand the decisions of the model.

Most current VLA models produce the two outputs in parallel (\cref{fig:teaser}c).
OmniDrive~\cite{omnidrive} lets the language head output both the language decision and the trajectory as text~\cite{gptdriver,drivegpt4,drivevlm,emma}.
Text tokens are discrete and carry no notion of numerical distance, so waypoints generated token by token are not precise enough~\cite{orion}.
ORION~\cite{orion} improves on this design and predicts the waypoints with a separate numerical planner~\cite{simlingo,recogdrive,alpamayor1}.
Combined with Reinforcement Learning (RL) post-training on the trajectory~\cite{recogdrive,minddrive}, this design alleviates the above problem to some extent.
However, neither design constrains the two outputs, which are often inconsistent~\cite{vlafaithful,couplingcf,drivethoughts}.
On Bench2Drive~\cite{bench2drive}, the decision stated by MindDrive contradicts its own planned trajectory on the vast majority of the frames (\cref{fig:teaser}a and \cref{tab:gap}).
This is because the language annotations of the training data are not checked against the expert trajectory, and no explicit constraint links the decision to the trajectory.
The trajectory is usually the more accurate of the two (\cref{sec:exp:gap}), because subsequent training, including RL post-training, mainly rewards the trajectory and neglects the language decision.
As a result, the language decision becomes redundant, and the interpretability that the VLA paradigm promises is greatly limited.
Recent methods such as MindDrive~\cite{minddrive} and DriveMA~\cite{drivema} attempt to solve this problem, but they align the two outputs only in form and do not effectively constrain their consistency.
The decision token of MindDrive selects among candidate trajectories that fall into the same maneuver class on most frames, so it is an index without a meaning (\cref{fig:teaser}b).
DriveMA considers the future motion of the expert only in the labels of the language decision and does not resolve the contradictions inside the chain-of-thought reasoning.
A data pipeline that enforces the consistency of the whole annotation is still missing.

To address this challenge, we propose \method, a VLA framework that imposes the language decision as a consistency constraint on trajectory planning (\cref{fig:teaser}c).
Specifically, the language decision is stated as a \emph{commitment}, which consists of a maneuver class, such as slow down, and a target value, such as the speed to reach.
Facts reported by the perception of the model rule out the decisions that the scene does not allow.
A \codebook then maps the commitment to a code with a kinematic meaning, and the planner generates the trajectory conditioned on this code.
To make the planner follow the decision, we design consistency losses that supervise every admissible decision of a scene, with the expert trajectory for the decision of the expert and a counterfactual execution for each alternative.
To provide the supervision, we build an automatic data pipeline in which rule-based functions and a VLM generate, review and refine the language annotations.
In addition, the mapping takes only a decision as input, so the framework is extensible to mature rule-based decision modules, which can be plugged in without modification.
Extensive experiments on the closed-loop Bench2Drive benchmark show that \method, with a lightweight 0.5B language model, is competitive with the state of the art, reaching a Driving Score (DS) of 78.50 and a Success Rate (SR) of 55.91\%, and reduces the inconsistency between the language decision and the planned trajectory to 1.8\%.

Our main contributions are summarized as follows:
\begin{itemize}
\item We propose \method, a language-constrained planning framework for VLA autonomous driving models. By introducing a \codebook and consistency losses, \method turns the language decision from a mode index into a semantic constraint that the trajectory is trained to satisfy, substantially improving the consistency between language decisions and planned trajectories.
\item We introduce an automatic data pipeline that combines rule-based functions with a VLM to generate, review and refine language annotations, reducing their reasoning contradictions. To the best of our knowledge, it is the first to ground counterfactual reasoning in the facts of visual perception and to train the language decision and the trajectory on it together, so that the two outputs of the model stay consistent.
\item Extensive experiments demonstrate the effectiveness of \method, achieving 78.50 DS and 55.91\% SR on the Bench2Drive benchmark with a lightweight 0.5B language model. The lookup of the \codebook also decouples the language decision from the hidden states of the language model, which makes the framework extensible to rule-based decisions.
\end{itemize}

\begin{figure*}[t]
\centering
\includegraphics[width=\linewidth]{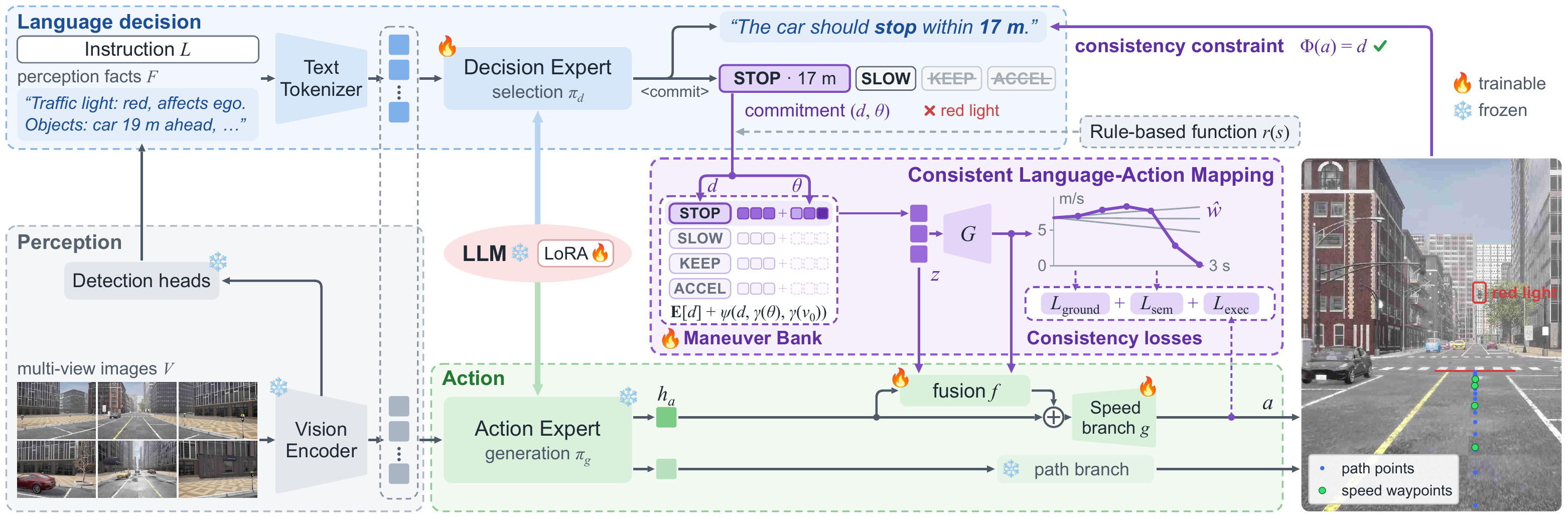}
\caption{\textbf{Overview of \method.} From the perception facts $F$ and the instruction $L$, the Decision Expert ($\pi_d$) outputs the language decision as a commitment $(d,\theta)$ inside the admissible set. The \codebook encodes $(d,\theta)$ into the code $z$ and decodes a nominal speed profile $\hat w$; both are fused into the speed branch of the Action Expert ($\pi_g$), whose trajectory $a$ is trained to satisfy the consistency constraint $\Phi(a)=d$. Consistency losses train this mapping, and a rule-based function $r(s)$ can issue the commitment.}
\label{fig:framework}
\end{figure*}

\section{Related Work}
\label{sec:related}
We review E2E driving and language-action consistency.

\subsection{End-to-End Autonomous Driving}
E2E autonomous driving learns perception, prediction and planning in one differentiable network that maps sensor inputs to trajectories~\cite{transfuser,tcp,stp3,uniad,vad}.
UniAD~\cite{uniad} and VAD~\cite{vad} unify several tasks on bird's-eye-view representations, and GenAD~\cite{genad} and DiffAD~\cite{diffad} model planning as a generative problem.
However, these models lack the reasoning ability and common sense that complex scenarios require, so recent methods build driving models on VLMs~\cite{lmdrive,opendrivevla,spacedrive,adathinkdrive}.
DriveGPT4~\cite{drivegpt4}, EMMA~\cite{emma} and OmniDrive~\cite{omnidrive} output control signals or trajectories as text.
Senna~\cite{senna} and DriveVLM~\cite{drivevlm} pair a VLM with a separate planner, and ORION~\cite{orion} links the VLM to a generative planner with a planning token.
AutoVLA~\cite{autovla}, ReCogDrive~\cite{recogdrive} and AlphaDrive~\cite{alphadrive} are further optimized with RL.
MindDrive~\cite{minddrive} uses a Decision Expert and an Action Expert that share one language model, and optimizes its language decisions with trajectory-level rewards in online RL.

\subsection{Language-Action Consistency}
The explanation of a model need not reflect the computation behind its prediction~\cite{jacovi2020faithful,turpin2023unfaithful,lanham2023faithful}.
In driving, prior works align textual explanations or reasoning with the predicted actions~\cite{bddx,adapt,rdadriver}.
SimLingo~\cite{simlingo} trains the planner on simulated alternative futures to follow language instructions.
Alpamayo-R1~\cite{alpamayor1} rewards the consistency between reasoning and action in RL post-training, and CF-VLA~\cite{cfvla} revises its meta-actions by counterfactual reflection.
DriveMA~\cite{drivema} labels meta-actions from the future motion of the expert, generates the meta-action before the trajectory and rewards their agreement.
Audits find the reasoning of driving VLMs and VLA models often ungrounded or weakly coupled to the trajectory~\cite{drivebench,vlafaithful,couplingcf}, and run-time monitors flag or reject such mismatches~\cite{drivethoughts,seal}.
These methods measure or encourage consistency through data, losses and rewards, while guaranteed constraints rely mainly on rule-based executors and shields outside VLA planners~\cite{drivemlm,languagempc,alshiekh2018shielding,rss,vlsa}.

\section{Method}
\label{sec:method}

In this section, we present \method in detail.
As illustrated in \cref{fig:framework}, following MindDrive~\cite{minddrive}, \method comprises a Decision Expert and an Action Expert, which share one VLM and differ only in their Low-Rank Adaptation (LoRA)~\cite{lora} parameters.
The Decision Expert serves as the language decision head and states the driving decision, while the Action Expert serves as the trajectory execution head and generates the trajectory.
We first formulate trajectory planning under a language consistency constraint (\cref{sec:method:problem}).
We then introduce the consistent language-action mapping, which maps the language decision to a code with a kinematic meaning through a \codebook and trains the Action Expert with consistency losses to follow it (\cref{sec:method:mapping}).
Next, the counterfactual decision-action data pipeline pairs every admissible decision of a scene with a trajectory, and reviews and refines the language annotations (\cref{sec:method:data}).
Finally, the mapping makes \method extensible to rule-based decisions, which drive the same Action Expert (\cref{sec:method:rule}).

\subsection{Problem Formulation}
\label{sec:method:problem}

We formulate end-to-end autonomous driving as generating a set of $N$ candidate trajectories $A$ and determining the optimal trajectory $a^*$ from the surrounding visual information $V$ and the language instruction $L$:
\begin{equation}
\label{eq:task}
\begin{aligned}
A&=\{a_i\}_{i=1}^{N},\quad a_i\sim\pi_g(a\mid V,L),\\
a^*&=\argmax_{a\in A}\,\pi_g(a\mid V,L),
\end{aligned}
\end{equation}
where $a_i$ is a trajectory of future waypoints and $\pi_g$ is the trajectory generation policy.
MindDrive~\cite{minddrive} further models the selection of $a^*$ as a decision process of the VLA model and introduces a selection policy $\pi_d$:
\begin{equation}
\label{eq:select}
a^*=\argmax_{a\in A}\,\underbrace{\pi_d(a\mid V,L)}_{\text{selection}}\cdot\underbrace{\pi_g(a\mid V,L)}_{\text{generation}}.
\end{equation}
\label{eq:implicit}%
From \cref{eq:select}, the optimal trajectory depends on two policies, $\pi_d$ and $\pi_g$.
However, the two policies are connected only by an index. The language decision selects one candidate, and no term requires this candidate to follow the decision.
To solve this problem, we make the language decision $d$ explicit and insert a consistency constraint between the two policies:
\begin{equation}
\label{eq:hard}
\begin{aligned}
a^*=\argmax_{a\in A,\,d\in\Dset}\;&\underbrace{\pi_d(d\mid V,L)}_{\text{selection}}\cdot\underbrace{\mathbb{I}\big[\Phi(a)=d\big]}_{\text{consistency constraint}}\\
&\cdot\underbrace{\pi_g(a\mid V,L,d)}_{\text{generation}},
\end{aligned}
\end{equation}
where $d\in\Dset$ is the language decision, $\Dset$ is the decision space, $\mathbb{I}[\cdot]$ is the indicator function, $\Phi(a)$ denotes the decision that the trajectory $a$ realizes, and the generation policy $\pi_g$ is conditioned on $d$.
The decision $d$, which abbreviates the commitment of \cref{sec:method:mapping}, is selected first, and the trajectory is generated under it (\cref{eq:rule}).
We relax the indicator into the consistency losses of \cref{sec:method:mapping} and do not enforce it at inference, so the constraint is learned and not guaranteed (\cref{sec:exp:gap}).

\subsection{Consistent Language-Action Mapping}
\label{sec:method:mapping}
\label{sec:method:do}%

To satisfy \cref{eq:hard}, the language decision and the action must be consistent, and the language decision must constrain the action directly.
This requires the language decision to carry semantics that the action can follow.
However, in existing methods the decision reaches the action only as an index, which carries no semantics.
We therefore introduce a \codebook, which gives the language decision a kinematic meaning and maps it to the action, and consistency losses, which train the Action Expert to follow it.

\paragraph{\codebook.}
The Decision Expert implements the selection policy $\pi_d$.
Its prompt carries a set of perception facts $F$, short sentences rendered from the detection heads of the model, which define the admissible set $\Dadm\subseteq\Dset$ of the decisions that are safe and legal in the scene.
At a special token \texttt{<commit>}, the Decision Expert states the decision as a sentence and outputs it as a \emph{commitment} $c=(d,\theta)$, a parameterized action~\cite{masson2016parameterized,hausknecht2016parameterized}, where $d\in\Dadm$ is one of four speed decisions, \textsc{stop}, \textsc{slow}, \textsc{keep} and \textsc{accel}, and $\theta$ is its target value, the stopping distance for \textsc{stop} and the end speed otherwise.
Both forms are supervised by the same label, which is read from the expert trajectory.

The \codebook then maps the commitment to the action.
It stores one learnable entry for each decision, and the decision $d$ looks up its entry.
The target value $\theta$ and the current speed $v_0$ are encoded with Fourier features~\cite{tancik2020fourier}, passed through $\psi$ together with $d$ and added to the entry, which forms the commitment code $z$.
A grounding head decodes $z$ into a nominal speed profile $\hat w$, and both are fused into the speed branch of the Action Expert, which implements the generation policy $\pi_g$.
The \codebook is summarized as:
\begin{equation}
\label{eq:code}
\begin{aligned}
\text{input:}\quad & c=(d,\theta),\;\; v_0,\\
\text{lookup:}\quad & z=\mathbf{E}[d]+\psi\big(d,\gamma(\theta),\gamma(v_0)\big),\\
\text{output:}\quad & \hat w=G(z,v_0),\\
& a=g\big(h_a+f([h_a;z;\hat w;v_0])\big),
\end{aligned}
\end{equation}
where $\mathbf{E}[d]$ is the entry of decision $d$, $\gamma$ denotes the Fourier features, $\psi$ and $f$ are Multi-Layer Perceptrons (MLPs), $G$ is the grounding head, $h_a$ is the hidden state of the Action Expert, $g$ is its speed branch, and $a$ is the generated trajectory, one per admissible decision in the set $A$ of \cref{eq:task}.
The lookup involves no hidden state of the language model, so the Action Expert receives exactly what the decision says.

\paragraph{Consistency losses.}
A conditioned planner can still ignore its condition. On expert data, the decision is a function of the scene, so the planner can reproduce the expert from $h_a$ alone.
We therefore supervise every admissible decision of a scene.
For each $k\in\Dadm$, the target $\bar a_k$ is the expert trajectory if $k$ is the decision of the expert, and a counterfactual execution of $k$ in the same scene otherwise (\cref{sec:method:data}).
Given the commitment $(k,\theta_k)$ read from $\bar a_k$, the \codebook outputs the profile $\hat w_k$ and the Action Expert generates $a_k$.
We define three consistency losses:
\begin{equation}
\label{eq:cons}
\begin{aligned}
\mathcal{L}_{\mathrm{ground}}&=\big\|\hat w_k-w(\bar a_k)\big\|_1,\\
\mathcal{L}_{\mathrm{sem}}&=\big[m-\rho_k(\hat w_k)\big]_+,\\
\mathcal{L}_{\mathrm{exec}}&=\lambda_p\big\|a_k-\bar a_k\big\|_1+\big[m-\rho_k(w(a_k))\big]_+\\
&\quad+\big|\Theta(a_k)-\theta_k\big|,
\end{aligned}
\end{equation}
where $w(\cdot)$ is the speed profile of a trajectory, $\rho_k(\cdot)$ is the signed distance of a speed profile to the boundary of decision $k$ under $\Phi$, $[x]_+=\max(x,0)$, $m$ is a margin, $\Theta(\cdot)$ reads the target value of a trajectory, and $\lambda_p$ is a weight.
The grounding loss $\mathcal{L}_{\mathrm{ground}}$ ties the code to the motion that executes it.
The semantic loss $\mathcal{L}_{\mathrm{sem}}$ requires the decoded profile to be read back as decision $k$.
The execution loss $\mathcal{L}_{\mathrm{exec}}$ teaches the Action Expert what to generate under decision $k$, requires the output to be read back as $k$ with a hinge that relaxes the indicator of \cref{eq:hard}, and requires its target value to be the commanded one.
The same scene now has different targets under different decisions, and only the decision tells them apart.
The total consistency loss is:
\begin{equation}
\label{eq:consall}
\mathcal{L}_{\mathrm{cons}}=\frac{1}{|\Dadm|}\sum_{k\in\Dadm}\big(\mathcal{L}_{\mathrm{ground}}+\mathcal{L}_{\mathrm{sem}}+\mathcal{L}_{\mathrm{exec}}\big).
\end{equation}

\subsection{Counterfactual Decision-Action Data Pipeline}
\label{sec:method:data}
\label{sec:method:training}%
\begin{figure}[t]
\centering
\includegraphics[width=\linewidth]{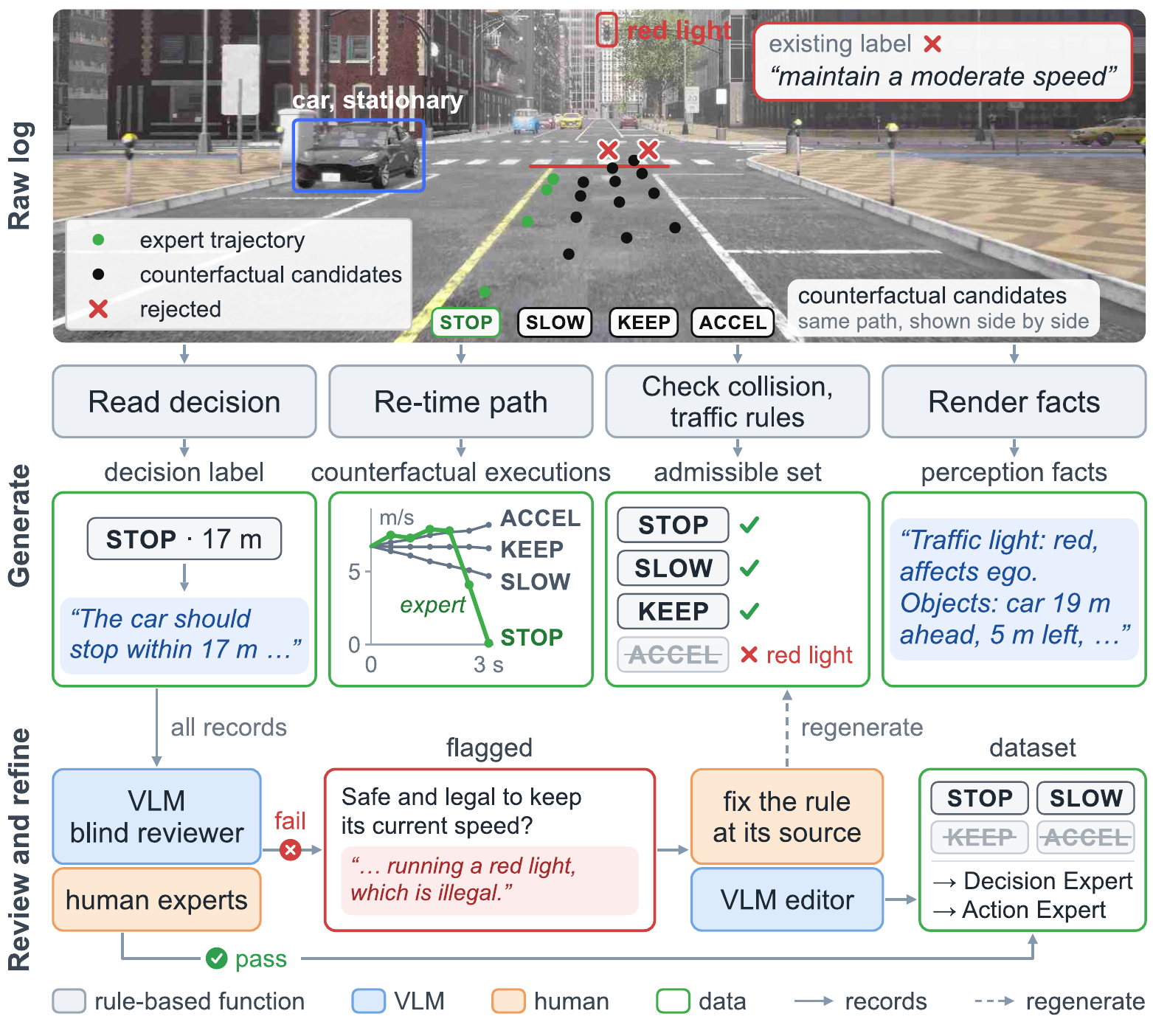}
\caption{\textbf{Counterfactual decision-action data pipeline.} From a raw log frame, rule-based functions \emph{generate} the decision label, the counterfactual executions, the admissible set and the perception facts. A VLM and human experts \emph{review} them; flagged records are fixed at the rule and regenerated, or \emph{refined}.}
\label{fig:pipeline}
\end{figure}

The mapping above needs supervision that existing datasets do not provide.
Their language annotations are not clearly tied to the action. A decision label often does not describe what the expert does next, and the reasoning text can contradict the scene or itself.
Moreover, counterfactual data has not yet been used to train the language decision and the action together.
To fill this gap, we build a counterfactual decision-action data pipeline.
As shown in \cref{fig:pipeline}, rule-based functions generate the records of every frame, and a VLM reviews and refines the language annotations.
The pipeline runs automatically on the raw logs of Bench2Drive~\cite{bench2drive}.

\paragraph{Generate.}
With one decision and one trajectory per scene, a model can produce both from the scene alone, without the trajectory following the decision.
We therefore pair every admissible decision of a scene with a trajectory, and rule-based functions write four records for every frame.
(i) \emph{Decision label}: the commitment $(\bar d,\bar\theta)$ is read from the next 3\,s of the expert trajectory and rendered as the decision sentence, so the label and the trajectory agree by construction.
(ii) \emph{Counterfactual executions}: for every other decision $k$, the path of the expert is re-timed with a speed profile of decision $k$, which gives a trajectory $\bar a_k$ with $\Phi(\bar a_k)=k$.
It is rolled out against the logged motion of the other agents on an occupancy grid~\cite{stp3} and marked infeasible if it collides.
(iii) \emph{Admissible set}: $\Dadm$ keeps the decisions whose execution is feasible and that no traffic rule forbids.
\label{eq:allowed}%
\textsc{keep} and \textsc{accel} are forbidden when the stop line of a red light or of a stop sign is within reach.
(iv) \emph{Perception facts}: the fact sentences, rendered from the logged scene by the templates that are used at inference.
With these records, the admissible set tells the Decision Expert which decisions are admissible, and the counterfactual executions tell the Action Expert what each of them looks like (\cref{eq:cons}).

\paragraph{Review and refine.}
A language annotation is a short chain of reasoning, and its contradictions are also learned by the model.
A VLM~\cite{qwen3vl} therefore takes two roles.
As a blind reviewer, it judges whether a decision sentence contradicts the scene, whether a fact sentence invents or omits an object, and whether an alternative decision is safe and legal.
As an editor, it revises reasoning text that contradicts itself.
Errors that originate in a rule or a template are corrected at their source, and the records are regenerated and reviewed again.
Finally, human experts check a random sample of the records.
The resulting dataset covers the base set of Bench2Drive (\SuppData).

\subsection{Extensibility to Rule-Based Decisions}
\label{sec:method:rule}

The mapping of \cref{sec:method:mapping} connects a coarse-grained decision to a fine-grained action.
The Action Expert reads the decision only through the code of \cref{eq:code}, which involves no hidden state of the language model.
The language decision and the action are therefore separable, and the Action Expert does not depend on the source of the decision.
Mature rule-based decision modules, such as traffic-rule logic, safety rules~\cite{rss} and rule-based planners~\cite{pdm}, already decide at this granularity.
Their output can be written as a commitment and executed by the same Action Expert without modification or retraining:
\begin{equation}
\label{eq:rule}
\begin{aligned}
a^*&=\argmax_{a\in A}\;\mathbb{I}\big[\Phi(a)=d\big]\cdot\pi_g(a\mid V,L,d),\\
d&=\begin{cases}
\argmax_{k\in\Dadm}\pi_d(k\mid V,L), & \text{(language)}\\
r(s), & \text{(rule)}
\end{cases},
\end{aligned}
\end{equation}
where the first line is the action side of \cref{eq:hard}, which depends on the decision only through $d$, $r$ is a rule-based function, and $s$ is the scene state that it reads from perception, such as the state of the traffic light or the time to collision.
The two cases give the source of $d$: the Decision Expert, or the rule-based function whenever it issues a decision.
Rules can therefore be added or exchanged without any change to the model.

\section{Experiments}
\label{sec:exp}
We evaluate \method and ablate its components.

\subsection{Experimental Settings}
\label{sec:exp:setting}
We detail the dataset, metrics, model and training.

\paragraph{Dataset.}
We train and evaluate \method on Bench2Drive~\cite{bench2drive}, a closed-loop benchmark for end-to-end autonomous driving based on the CARLA simulator~\cite{carla}.
We adopt the official base set of 1000 clips, with 950 for training and 50 for open-loop validation.
The clips are annotated by our data pipeline (\cref{sec:method:data}), which yields 234{,}769 training and 12{,}806 validation frames.
Closed-loop evaluation uses the official 220 short routes across 44 interactive scenarios.

\paragraph{Metrics.}
For closed-loop evaluation, we adopt DS, SR and Multi-Ability.
Following Bench2Drive~\cite{bench2drive}, SR is the proportion of routes completed within the time limit, DS is the route completion penalized by traffic infractions, and Multi-Ability evaluates five advanced urban driving skills.
To measure the consistency between the language decision and the trajectory, we further report the say-do gap:
\begin{equation}
\label{eq:saydo}
\mathrm{SD}=\frac{1}{N_f}\sum_{t=1}^{N_f}\mathbb{I}\big[\Phi(a_t)\neq d_t\big],
\end{equation}
where $N_f$ is the number of frames, $d_t$ is the decision stated at frame $t$ and $a_t$ is the planned trajectory.
The say-exec gap replaces $a_t$ by the executed trajectory.

\paragraph{Model Setting.}
We set the number of speed waypoints to $N_s=6$, a future 3\,s trajectory sampled at 2\,Hz.
The decision space of the Decision Expert includes 4 speed decisions, each with a continuous target value.
\label{eq:phi}%
$\Phi$ reads a trajectory as \textsc{stop} if its end speed is below 0.5\,m/s, as \textsc{slow} or \textsc{accel} if its speed changes by more than 1\,m/s, and as \textsc{keep} otherwise.
The \codebook has 4 entries of dimension 64 and uses Fourier features with 4 frequencies, and its grounding head decodes $N_s$ speed values.
The fusion MLP has a hidden size of 512 and a zero-initialized output layer.
In the consistency losses, we set $m=0.3$\,m/s and $\lambda_p=3$.

\paragraph{Training Process.}
All models are trained on 4 NVIDIA A100 graphics cards.
We employ the pretrained EVA-02-L~\cite{eva02} as the vision encoder and the lightweight Qwen2-0.5B~\cite{qwen2} as the base Large Language Model (LLM).
\method is trained in two stages, Imitation Learning (IL) and online RL, both following MindDrive~\cite{minddrive}.
Only the Action Expert is initialized from the released checkpoint of MindDrive, which gives the trajectory generation a prior.
This Action Expert has been trained by IL alone, as the online RL of MindDrive updates only its Decision Expert and value network.
In the IL stage, the Decision Expert is fine-tuned with LoRA to state the decision, and the \codebook, the fusion MLP and the speed branch are then trained with $\mathcal{L}_{\mathrm{cons}}$, while all other modules are frozen.
Closed-loop evaluation runs online on a single RTX 4090, and every measured result is the median over multiple random seeds.

\subsection{Main Results}
\label{sec:exp:main}
\begin{table*}[t]
\centering
\caption{\textbf{Closed-loop and multi-ability results on Bench2Drive}, in the protocol and layout of MindDrive~\cite{minddrive}. C/L: camera/LiDAR. $^*$: expert feature distillation. $^\S$: trained on its own dataset instead of the base set. DS: driving score; SR: success rate; M: merging; O: overtaking; EB: emergency brake; GW: give way; TS: traffic sign. Avg.\ L2 is averaged over 2\,s at 2\,Hz as in UniAD~\cite{uniad}. Upper blocks: as published. Best among the methods trained on the base set in bold. $^\ddagger$: evaluated by us on a single RTX 4090 with the official evaluator under one simulator build, median over multiple random seeds; the baselines are the released checkpoints, with reproduced numbers in italics. $^\dagger$: Action Expert initialized from the released MindDrive checkpoint, in which it is trained by IL only.}
\label{tab:main}
\small
\setlength{\tabcolsep}{4pt}
\renewcommand{\arraystretch}{0.9}%
\resizebox{\textwidth}{!}{%
\begin{tabular}{@{}lllccccccccccc@{}}
\toprule
\multirow{2}{*}{Method} & \multirow{2}{*}{LLM} & \multirow{2}{*}{Reference} & \multirow{2}{*}{Modality} & \multirow{2}{*}{Scheme} & \multicolumn{2}{c}{Closed-loop} & \multicolumn{6}{c}{Ability (\%)\,$\uparrow$} & Open-loop \\
\cmidrule(lr){6-7}\cmidrule(lr){8-13}\cmidrule(l){14-14}
 & & & & & DS\,$\uparrow$ & SR (\%)\,$\uparrow$ & M & O & EB & GW & TS & Mean & Avg.\ L2\,$\downarrow$ \\
\midrule
\multicolumn{14}{@{}l}{\emph{Traditional end-to-end paradigm}} \\
TCP-traj$^*$~\cite{tcp} & -- & NeurIPS 22 & C & IL & 59.90 & 30.00 & 8.89 & 24.29 & 51.67 & 40.00 & 46.28 & 34.22 & 1.70 \\
ThinkTwice$^*$~\cite{thinktwice} & -- & CVPR 23 & C & IL & 62.44 & 31.23 & 27.38 & 18.42 & 35.82 & 50.00 & 54.23 & 37.17 & 0.95 \\
DriveAdapter$^*$~\cite{driveadapter} & -- & ICCV 23 & C\&L & IL & 64.22 & 33.08 & 28.82 & 26.38 & 48.76 & 50.00 & 56.43 & 42.08 & 1.01 \\
UniAD-Base~\cite{uniad} & -- & CVPR 23 & C & IL & 45.81 & 16.36 & 14.10 & 17.78 & 21.67 & 10.00 & 14.21 & 15.55 & 0.73 \\
VAD~\cite{vad} & -- & ICCV 23 & C & IL & 42.35 & 15.00 & 8.11 & 24.44 & 18.64 & 20.00 & 19.15 & 18.07 & 0.91 \\
GenAD~\cite{genad} & -- & ECCV 24 & C & IL & 44.81 & 15.90 & -- & -- & -- & -- & -- & -- & -- \\
MomAD~\cite{momad} & -- & CVPR 25 & C & IL & 44.54 & 16.71 & -- & -- & -- & -- & -- & -- & 0.87 \\
WoTE~\cite{wote} & -- & ICCV 25 & C\&L & IL & 61.71 & 31.36 & -- & -- & -- & -- & -- & -- & -- \\
DriveTransformer-L~\cite{drivetransformer} & -- & ICLR 25 & C & IL & 63.46 & 35.01 & 17.57 & 35.00 & 48.36 & 40.00 & 52.10 & 38.60 & 0.62 \\
DiffAD~\cite{diffad} & -- & arXiv 25 & C & IL & 67.92 & 38.64 & 30.00 & 35.55 & 46.66 & 40.00 & 46.32 & 38.79 & 1.55 \\
PGS~\cite{pgs} & -- & NeurIPS 25 & C & IL & 78.08 & 48.64 & 35.00 & 73.33 & 55.00 & \textbf{60.00} & 43.68 & 53.40 & 0.77 \\
DriveDPO~\cite{drivedpo} & -- & NeurIPS 25 & C & Offline RL & 62.02 & 30.62 & -- & -- & -- & -- & -- & -- & -- \\
Raw2Drive~\cite{raw2drive} & -- & NeurIPS 25 & C & Online RL & 71.36 & 50.24 & \textbf{43.35} & 51.11 & 60.00 & 50.00 & 62.26 & 53.34 & -- \\
\midrule
\multicolumn{14}{@{}l}{\emph{Vision-language-action paradigm}} \\
Drive-$\pi_0$~\cite{drivemoe} & PaliGemma-3B & arXiv 25 & C & IL & 60.45 & 30.00 & 29.35 & 36.58 & 48.83 & 40.00 & 54.45 & 41.84 & 0.56 \\
DriveMoE~\cite{drivemoe} & PaliGemma-3B & arXiv 25 & C & IL & 74.22 & 48.64 & 34.67 & 40.00 & 65.45 & 40.00 & 59.44 & 47.91 & \textbf{0.38} \\
ORION~\cite{orion} & Vicuna-7B & ICCV 25 & C & IL & 77.74 & 54.62 & 25.00 & 71.11 & 78.33 & 30.00 & 69.15 & 54.72 & 0.68 \\
ORION~\cite{orion} & Qwen2-0.5B & ICCV 25 & C & IL & 72.89 & 45.83 & 26.32 & 62.22 & 55.55 & 50.00 & 63.30 & 51.37 & 0.67 \\
UniDrive-WM~\cite{unidrivewm} & Vicuna-7B & arXiv 26 & C & IL & \textbf{79.22} & 56.36 & 29.81 & 74.04 & \textbf{79.84} & 40.00 & 71.30 & 59.01 & 0.64 \\
ReCogDrive~\cite{recogdrive} & Qwen2.5-7B & ICLR 26 & C & Offline RL & 71.36 & 45.45 & 29.73 & 20.00 & 69.09 & 20.00 & \textbf{71.34} & 42.03 & -- \\
MindDriver~\cite{minddriver} & Qwen2.5-3B & CVPR 26 & C & Offline RL & 65.48 & 39.55 & -- & -- & -- & -- & -- & -- & -- \\
AutoVLA~\cite{autovla} & Qwen2.5-3B & NeurIPS 25 & C & Offline RL & 78.84 & \textbf{57.73} & -- & -- & -- & -- & -- & -- & -- \\
SimLingo$^\S$~\cite{simlingo} & InternVL2-1B & CVPR 25 & C & IL & 85.07 & 67.27 & -- & -- & -- & -- & -- & -- & -- \\
MindDrive~\cite{minddrive} & Qwen2-0.5B & ECCV 26 & C & IL & 75.85 & 49.30 & 30.26 & 66.67 & 60.00 & 33.33 & 56.91 & 49.44 & 0.69 \\
MindDrive~\cite{minddrive} & Qwen2-0.5B & ECCV 26 & C & Online RL & 78.04 & 55.09 & 32.89 & \textbf{75.56} & 68.33 & 50.00 & 57.89 & 56.94 & 0.69 \\
\midrule
\multicolumn{14}{@{}l}{\emph{Reproduced on a local RTX 4090}} \\
ORION$^\ddagger$ & Vicuna-7B & ICCV 25 & C & IL & \repro{75.40} & \repro{49.55} & \repro{22.78} & \repro{57.78} & \repro{73.33} & \repro{40.00} & \repro{65.26} & \repro{51.83} & \repro{0.69} \\
MindDrive$^\ddagger$ & Qwen2-0.5B & ECCV 26 & C & Online RL & \repro{73.24} & \repro{46.59} & \repro{28.48} & \repro{60.00} & \repro{63.33} & \repro{35.00} & \repro{61.05} & \repro{49.57} & \repro{0.69} \\
\textbf{\method}$^{\dagger\ddagger}$ & Qwen2-0.5B & -- & C & IL & 76.50 & 50.00 & 30.38 & 66.67 & 61.67 & 40.00 & 57.37 & 51.22 & 0.685 \\
\textbf{\method}$^{\dagger\ddagger}$ & Qwen2-0.5B & -- & C & Online RL & 78.50 & 55.91 & 32.91 & \textbf{75.56} & 70.00 & \textbf{60.00} & 58.95 & \textbf{59.48} & 0.60 \\
\bottomrule
\end{tabular}}
\end{table*}

\begin{figure*}[t]
\centering
\includegraphics[width=\linewidth]{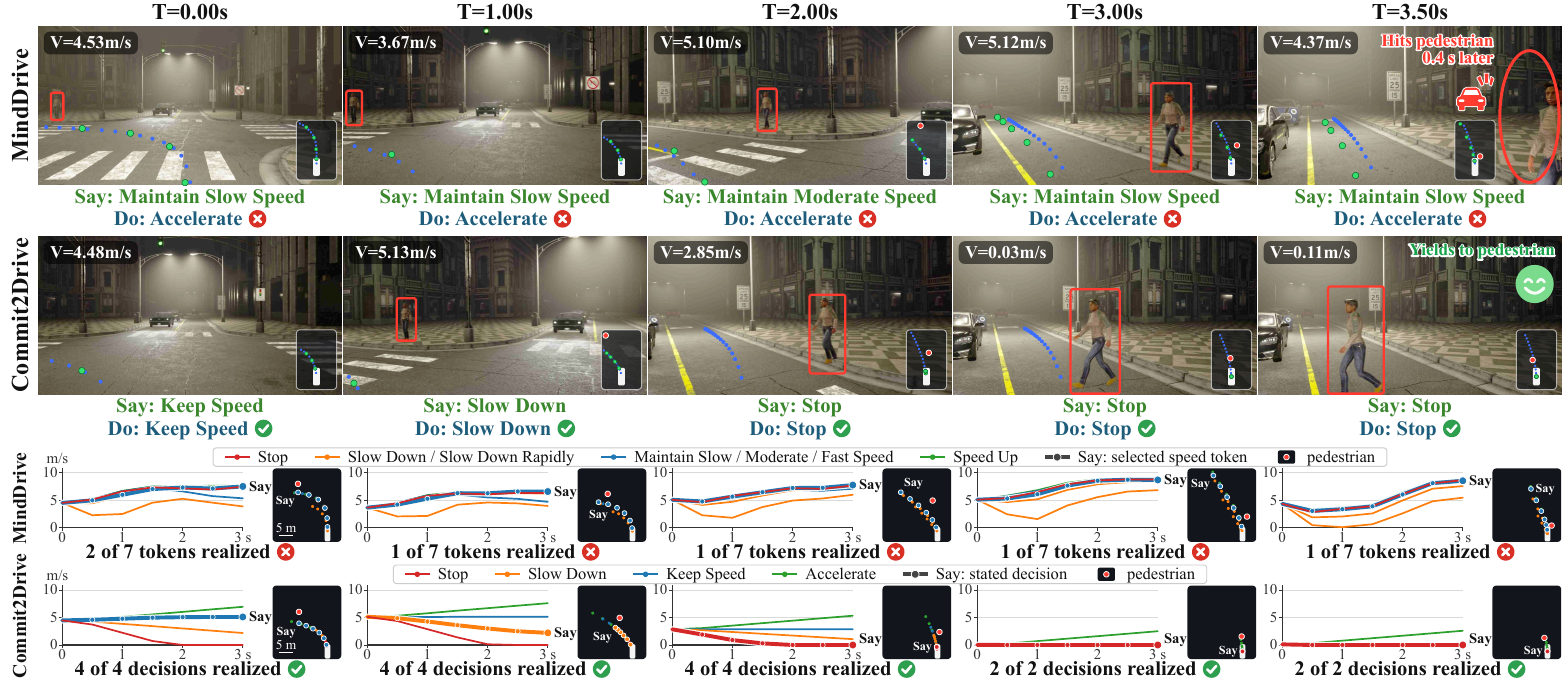}
\caption{\textbf{Qualitative comparison on Bench2Drive}, closed-loop route 11381 of the 220 official routes (scenario \textit{VehicleTurningRoutePedestrian}, Town12). Blue and green points are the predicted path points and speed waypoints. \textit{Say} is the stated decision and \textit{Do} the decision realized by the speed waypoints. MindDrive, from our log of its released checkpoint, hits the pedestrian 0.4\,s after the last frame. Bottom rows: speed over the next 3\,s and top view of the candidate trajectory of every speed token or decision; those of MindDrive are recomputed by replaying the logged frames.}
\label{fig:qualitative}
\end{figure*}

We compare \method with representative methods of the traditional E2E paradigm and the VLA paradigm on Bench2Drive.
\cref{tab:main} lists the results, and we find that:

\textbf{1) \method delivers competitive closed-loop and open-loop performance with a lightweight model.}
With the Qwen2-0.5B language model, \method achieves 76.50 DS and 50.00\% SR after IL, and 78.50 DS and 55.91\% SR after online RL.
Among the VLA models with the same lightweight LLM, it surpasses ORION~\cite{orion} (72.89 DS, 45.83\% SR) and is comparable to MindDrive~\cite{minddrive}, which reports 75.85 DS after IL and 78.04 DS after online RL.
SimLingo~\cite{simlingo} reaches 85.07 DS with a model of similar size, but it is trained on its own dataset and not on the base set.
Reproduced on our RTX 4090 ($^\ddagger$ in \cref{tab:main}), MindDrive reaches 73.24 DS and 46.59\% SR, and \method improves over it by 5.26 DS and 9.32\% SR.
In Multi-Ability, it achieves a mean of 59.48\%.
In open loop, \method obtains an average L2 error of 0.685\,m after IL, close to MindDrive (0.69\,m) and ORION with the same LLM (0.67\,m), and 0.60\,m after online RL.

\textbf{2) The language decision of \method is consistent with its trajectory.}
\label{sec:exp:gap}
\label{sec:exp:cause}
\begin{table}[t]
\centering
\caption{\textbf{Say-do consistency on Bench2Drive.} Say-do: \cref{eq:saydo}; say-exec: the same with the executed trajectory. Lon.: speed decision; Yield: yielding action (\textsc{stop} or \textsc{slow}). ORION outputs no speed token. \method: Dev10.}
\label{tab:gap}
\small
\setlength{\tabcolsep}{4pt}
\begin{tabular}{@{}lcccc@{}}
\toprule
 & \multicolumn{2}{c}{Say-do (\%)\,$\downarrow$} & \multicolumn{2}{c}{Say-exec (\%)\,$\downarrow$} \\
\cmidrule(lr){2-3}\cmidrule(l){4-5}
Method & Lon. & Yield & Lon. & Yield \\
\midrule
ORION~\cite{orion} & -- & 23.8 & -- & 65.7 \\
MindDrive~\cite{minddrive} & 83.5 & 19.8 & 69.6 & 59.9 \\
\textbf{\method} & 1.8\dif{\textminus 81.7} & 0.8\dif{\textminus 19.0} & 52.4\dif{\textminus 17.2} & 39.8\dif{\textminus 20.1} \\
\bottomrule
\end{tabular}
\end{table}

\cref{tab:gap} compares the say-do gap of \cref{eq:saydo}.
MindDrive and ORION are evaluated from their released checkpoints, and their stated decisions are mapped to $\Dset$ (\SuppData).
The decision stated by MindDrive differs from its planned trajectory on 83.5\% of the frames and from the executed trajectory on 69.6\%.
On the open-loop validation frames, where the expert is available, its say-do gap is 66.4\%, and the language decision is wrong in 71\% of these conflicts and the trajectory in 13\%.
In contrast, \method reduces the say-do gap to 1.8\% in closed loop on Dev10, a ten-route subset of Bench2Drive, and to 0.4\% on the open-loop validation frames.
With the decision forced to another value, the trajectory of \method follows it on at least 99\% of the frames, against 27\% to 30\% for MindDrive (\cref{fig:teaser}b), which shows that the decision determines the trajectory.
The say-exec gap of \method is 17.2 points below that of MindDrive but remains 52.4\%.
With a say-do gap of 1.8\%, this remainder is the gap between the plan of one step and the motion executed under re-planning, which the consistency constraint does not cover.
The say-exec gap of SimLingo is 21.0\% at a say-do gap of 49.3\% (\SuppAnalysis).

\subsection{Qualitative Results}
\label{sec:exp:qual}
\cref{fig:qualitative} compares MindDrive and \method on a pedestrian-crossing scenario of the Bench2Drive closed-loop evaluation.
MindDrive states that it maintains a slow or moderate speed in all five frames, while its speed waypoints accelerate, and it hits the pedestrian 0.4\,s after the last frame.
We further examine the semantics of the decision on a subset of 3,434 validation frames (\cref{fig:teaser}b).
The seven candidate trajectories behind the speed tokens of MindDrive are nearly identical. On 86\% of the frames they fall into the same decision, two candidates end only 0.51\,m apart on average, and the output follows its token on fewer than one third of them.
In \method, the four decisions of a scene yield distinct trajectories, 8.38\,m apart on average, and the output follows the stated decision on at least 99\% of the frames.

\begin{table}[t]
\centering
\caption{\textbf{Ablation of \method.} Each row removes one component: the \codebook (the decision is passed as an index), the counterfactual data (only the decision of the expert is supervised) or the consistency losses (\cref{eq:consall}). L2: open-loop error at 3\,s.}
\label{tab:ablation}
\small
\setlength{\tabcolsep}{4pt}
\resizebox{\linewidth}{!}{%
\begin{tabular}{@{}lcccc@{}}
\toprule
Variant & Say-do (\%)\,$\downarrow$ & L2 (m)\,$\downarrow$ & DS\,$\uparrow$ & SR (\%)\,$\uparrow$ \\
\midrule
\textbf{\method} & 2.0 & 1.19 & 76.5 & 50.00 \\
w/o \codebook & 15.0\dif{+13.0} & 1.22\dif{+0.03} & 75.2\dif{\textminus 1.3} & 48.18\dif{\textminus 1.82} \\
w/o counterfactual data & 25.0\dif{+23.0} & 1.20\dif{+0.01} & 73.6\dif{\textminus 2.9} & 46.82\dif{\textminus 3.18} \\
w/o consistency losses & 8.0\dif{+6.0} & 1.21\dif{+0.02} & 75.6\dif{\textminus 0.9} & 48.64\dif{\textminus 1.36} \\
\bottomrule
\end{tabular}}
\end{table}

\subsection{Ablation Studies}
\label{sec:exp:ablation}
We ablate \method and test its extensibility to rules.

\paragraph{Bank, data and consistency losses.}
\cref{tab:ablation} ablates the \codebook, the counterfactual data and the consistency losses of \method.
The counterfactual data matters most. With supervision on the decision of the expert only, the say-do gap rises from 2.0\% to 25.0\% and the DS falls by 2.9, so the Action Expert follows the decision less reliably.
Further ablations are listed in \SuppAblation.

\paragraph{Extensibility to rule-based decisions.}
\label{sec:exp:rule}\label{sec:exp:speed}
\begin{table}[t]
\centering
\caption{\textbf{Latency breakdown with rule-based decisions.} Median model time per step in ms on a single RTX 4090, without simulator and control, on the Dev10 subset of Bench2Drive. Perception: vision encoder and perception heads. Decision: output of the language decision. Planner: output of the trajectory.}
\label{tab:rule}
\label{tab:speed}%
\small
\setlength{\tabcolsep}{2.5pt}
\begin{tabular}{@{}lcccc@{}}
\toprule
Method & Perception\,$\downarrow$ & Decision\,$\downarrow$ & Planner\,$\downarrow$ & Total\,$\downarrow$ \\
\midrule
ORION~\cite{orion} & \repro{165.6} & \repro{79.6} & \repro{5.8} & \repro{251.1} \\
MindDrive~\cite{minddrive} & \repro{185.3} & \repro{34.5} & \repro{26.3} & \repro{246.2} \\
\textbf{\method} & 189.8 & 35.5 & 33.1 & 259.8 \\
\quad + rule-triggered & 177.0 & 5.4 & 33.0 & 216.7 \\
\bottomrule
\end{tabular}
\end{table}

We test the extensibility of \cref{sec:method:rule} by plugging two rule-based functions into \method without retraining.
A traffic-light rule decides \textsc{stop} or \textsc{slow} at a red light that governs the ego lane, and a collision-avoidance rule decides \textsc{stop} when a hazard is within the time-to-collision threshold.
A triggered rule issues the decision in place of the Decision Expert (\cref{eq:rule}), and the Action Expert is unchanged.
On Dev10, the DS changes from 53.09 to 51.57 with the rules active.
In \cref{tab:rule}, a rule-issued decision takes 5.4\,ms instead of 35.5\,ms, and perception accounts for most of the latency of every model.

\section{Conclusion}
\label{sec:conclusion}
In this paper, we introduce \method, a VLA framework that imposes the language decision as a consistency constraint on trajectory planning.
A \codebook maps the language decision to a code with a kinematic meaning, and consistency losses train the Action Expert to follow it.
An automatic data pipeline, in which rule-based functions and a VLM generate, review and refine the annotations, supplies the supervision.
Extensive experiments on Bench2Drive demonstrate that \method, with a lightweight 0.5B language model, is competitive with the state of the art and greatly reduces the inconsistency between the language decision and the planned trajectory.
The framework is also extensible to existing rule-based methods, whose decisions drive the same Action Expert without retraining.

\paragraph{Limitations and Future Work.}
This work addresses only the say-do gap inside the model, and its driving performance exceeds the baseline by a small margin.
The say-exec gap, which lies between the model and the physical execution of its plan, remains unsolved and is left to future work.
We hope that it brings new inspiration to the autonomous driving community on the consistency between language decisions and trajectories.

\section*{LLM Usage Statement}
\label{sec:llm}
We used an LLM-based assistant (Claude, Anthropic) to polish the language of the manuscript, to detect writing errors, to help draft the figures, and to help write and run code for the experiments.
The research idea, the method, the experiments and the writing of the paper are primarily the work of the authors, who checked all text, code and results and take full responsibility for the content.

{
    \small
    \bibliographystyle{ieeenat_fullname}
    \bibliography{main}

\begin{thebibliography}{64}
\providecommand{\natexlab}[1]{#1}
\providecommand{\url}[1]{\texttt{#1}}
\expandafter\ifx\csname urlstyle\endcsname\relax
  \providecommand{\doi}[1]{doi: #1}\else
  \providecommand{\doi}{doi: \begingroup \urlstyle{rm}\Url}\fi

\bibitem[Alshiekh et~al.(2018)Alshiekh, Bloem, Ehlers, K{\"o}nighofer, Niekum,
  and Topcu]{alshiekh2018shielding}
Mohammed Alshiekh, Roderick Bloem, R{\"u}diger Ehlers, Bettina K{\"o}nighofer,
  Scott Niekum, and Ufuk Topcu.
\newblock Safe reinforcement learning via shielding.
\newblock In \emph{AAAI}, 2018.

\bibitem[Bai et~al.(2025)]{qwen3vl}
Shuai Bai et~al.
\newblock {Qwen3-VL} technical report.
\newblock \emph{arXiv preprint arXiv:2511.21631}, 2025.

\bibitem[Chitta et~al.(2023)Chitta, Prakash, Jaeger, Yu, Renz, and
  Geiger]{transfuser}
Kashyap Chitta, Aditya Prakash, Bernhard Jaeger, Zehao Yu, Katrin Renz, and
  Andreas Geiger.
\newblock {TransFuser}: Imitation with transformer-based sensor fusion for
  autonomous driving.
\newblock \emph{IEEE Transactions on Pattern Analysis and Machine
  Intelligence}, 45\penalty0 (11):\penalty0 12878--12895, 2023.

\bibitem[Cohen(1960)]{cohen1960kappa}
Jacob Cohen.
\newblock A coefficient of agreement for nominal scales.
\newblock \emph{Educational and Psychological Measurement}, 20\penalty0
  (1):\penalty0 37--46, 1960.

\bibitem[Cui et~al.(2025)Cui, Wang, Li, Xie, Zou, Deng, Luo, Lu, Zhu, and
  Dai]{drivemlm}
Erfei Cui, Wenhai Wang, Zhiqi Li, Jiangwei Xie, Haoming Zou, Hanming Deng, Gen
  Luo, Lewei Lu, Xizhou Zhu, and Jifeng Dai.
\newblock {DriveMLM}: Aligning multi-modal large language models with
  behavioral planning states for autonomous driving.
\newblock \emph{Visual Intelligence}, 3:\penalty0 22, 2025.

\bibitem[Dauner et~al.(2023)Dauner, Hallgarten, Geiger, and Chitta]{pdm}
Daniel Dauner, Marcel Hallgarten, Andreas Geiger, and Kashyap Chitta.
\newblock Parting with misconceptions about learning-based vehicle motion
  planning.
\newblock In \emph{CoRL}, 2023.

\bibitem[Dosovitskiy et~al.(2017)Dosovitskiy, Ros, Codevilla, Lopez, and
  Koltun]{carla}
Alexey Dosovitskiy, German Ros, Felipe Codevilla, Antonio Lopez, and Vladlen
  Koltun.
\newblock {CARLA}: An open urban driving simulator.
\newblock In \emph{CoRL}, 2017.

\bibitem[Fang et~al.(2024)Fang, Sun, Wang, Huang, Wang, and Cao]{eva02}
Yuxin Fang, Quan Sun, Xinggang Wang, Tiejun Huang, Xinlong Wang, and Yue Cao.
\newblock {EVA-02}: A visual representation for neon genesis.
\newblock \emph{Image and Vision Computing}, 149:\penalty0 105171, 2024.

\bibitem[Fu et~al.(2025)Fu, Zhang, Zhao, Cui, Liang, Zhang, Zhang, Xie, Wang,
  and Bai]{orion}
Haoyu Fu, Diankun Zhang, Zongchuang Zhao, Jianfeng Cui, Dingkang Liang, Chong
  Zhang, Dingyuan Zhang, Hongwei Xie, Bing Wang, and Xiang Bai.
\newblock {ORION}: A holistic end-to-end autonomous driving framework by
  vision-language instructed action generation.
\newblock In \emph{Proceedings of the IEEE/CVF International Conference on
  Computer Vision}, pages 24823--24834, 2025.

\bibitem[Fu et~al.(2026)Fu, Zhang, Zhao, Cui, Xie, Wang, Chen, Ye, Liang, and
  Bai]{minddrive}
Haoyu Fu, Diankun Zhang, Zongchuang Zhao, Jianfeng Cui, Hongwei Xie, Bing Wang,
  Guang Chen, Hangjun Ye, Dingkang Liang, and Xiang Bai.
\newblock {MindDrive}: A vision-language-action model for autonomous driving
  via online reinforcement learning.
\newblock In \emph{ECCV}, 2026.
\newblock arXiv:2512.13636.

\bibitem[Hausknecht and Stone(2016)]{hausknecht2016parameterized}
Matthew Hausknecht and Peter Stone.
\newblock Deep reinforcement learning in parameterized action space.
\newblock In \emph{ICLR}, 2016.

\bibitem[Hu et~al.(2022{\natexlab{a}})Hu, Shen, Wallis, Allen-Zhu, Li, Wang,
  Wang, and Chen]{lora}
Edward~J. Hu, Yelong Shen, Phillip Wallis, Zeyuan Allen-Zhu, Yuanzhi Li, Shean
  Wang, Lu Wang, and Weizhu Chen.
\newblock {LoRA}: Low-rank adaptation of large language models.
\newblock In \emph{International Conference on Learning Representations
  (ICLR)}, 2022{\natexlab{a}}.

\bibitem[Hu et~al.(2022{\natexlab{b}})Hu, Chen, Wu, Li, Yan, and Tao]{stp3}
Shengchao Hu, Li Chen, Penghao Wu, Hongyang Li, Junchi Yan, and Dacheng Tao.
\newblock {ST-P3}: End-to-end vision-based autonomous driving via
  spatial-temporal feature learning.
\newblock In \emph{Computer Vision -- ECCV 2022}, pages 533--549,
  2022{\natexlab{b}}.

\bibitem[Hu et~al.(2026)Hu, Liu, Liu, Cen, Meng, Wang, Li, and He]{vlsa}
Songqiao Hu, Zeyi Liu, Shuang Liu, Jun Cen, Zihan Meng, Shihefeng Wang, Xiang
  Li, and Xiao He.
\newblock {VLSA}: Vision-language-action models with plug-and-play safety
  constraint layer.
\newblock In \emph{IROS}, 2026.

\bibitem[Hu et~al.(2023)Hu, Yang, Chen, Li, Sima, Zhu, Chai, Du, Lin, Wang, Lu,
  Jia, Liu, Dai, Qiao, and Li]{uniad}
Yihan Hu, Jiazhi Yang, Li Chen, Keyu Li, Chonghao Sima, Xizhou Zhu, Siqi Chai,
  Senyao Du, Tianwei Lin, Wenhai Wang, Lewei Lu, Xiaosong Jia, Qiang Liu,
  Jifeng Dai, Yu Qiao, and Hongyang Li.
\newblock Planning-oriented autonomous driving.
\newblock In \emph{Proceedings of the IEEE/CVF Conference on Computer Vision
  and Pattern Recognition (CVPR)}, pages 17853--17862, 2023.

\bibitem[Huang et~al.(2025)Huang, Qu, Jiang, Liu, and Zhang]{pgs}
Yi Huang, Zhan Qu, Lihui Jiang, Bingbing Liu, and Hongbo Zhang.
\newblock Prioritizing perception-guided self-supervision: A new paradigm for
  causal modeling in end-to-end autonomous driving.
\newblock In \emph{Advances in Neural Information Processing Systems
  (NeurIPS)}, 2025.

\bibitem[Huang et~al.(2024)Huang, Tang, Chen, Lin, Jie, Ma, Wang, and
  Liang]{rdadriver}
Zhijian Huang, Tao Tang, Shaoxiang Chen, Sihao Lin, Zequn Jie, Lin Ma, Guangrun
  Wang, and Xiaodan Liang.
\newblock Making large language models better planners with reasoning-decision
  alignment.
\newblock In \emph{Proc. European Conf. Computer Vision (ECCV)}, pages 73--90,
  2024.

\bibitem[Hwang et~al.(2025)Hwang, Xu, Lin, Hung, Ji, Choi, Huang, He,
  Covington, Sapp, Zhou, Guo, Anguelov, and Tan]{emma}
Jyh-Jing Hwang, Runsheng Xu, Hubert Lin, Wei-Chih Hung, Jingwei Ji, Kristy
  Choi, Di Huang, Tong He, Paul Covington, Benjamin Sapp, Yin Zhou, James Guo,
  Dragomir Anguelov, and Mingxing Tan.
\newblock {EMMA}: End-to-end multimodal model for autonomous driving.
\newblock \emph{Transactions on Machine Learning Research}, 2025.

\bibitem[Jacovi and Goldberg(2020)]{jacovi2020faithful}
Alon Jacovi and Yoav Goldberg.
\newblock Towards faithfully interpretable {NLP} systems: How should we define
  and evaluate faithfulness?
\newblock In \emph{ACL}, 2020.

\bibitem[Jia et~al.(2023{\natexlab{a}})Jia, Gao, Chen, Yan, Liu, and
  Li]{driveadapter}
Xiaosong Jia, Yulu Gao, Li Chen, Junchi Yan, Patrick~Langechuan Liu, and
  Hongyang Li.
\newblock {DriveAdapter}: Breaking the coupling barrier of perception and
  planning in end-to-end autonomous driving.
\newblock In \emph{ICCV}, 2023{\natexlab{a}}.

\bibitem[Jia et~al.(2023{\natexlab{b}})Jia, Wu, Chen, Xie, He, Yan, and
  Li]{thinktwice}
Xiaosong Jia, Penghao Wu, Li Chen, Jiangwei Xie, Conghui He, Junchi Yan, and
  Hongyang Li.
\newblock Think twice before driving: Towards scalable decoders for end-to-end
  autonomous driving.
\newblock In \emph{CVPR}, 2023{\natexlab{b}}.

\bibitem[Jia et~al.(2024)Jia, Yang, Li, Zhang, and Yan]{bench2drive}
Xiaosong Jia, Zhenjie Yang, Qifeng Li, Zhiyuan Zhang, and Junchi Yan.
\newblock {Bench2Drive}: Towards multi-ability benchmarking of closed-loop
  end-to-end autonomous driving.
\newblock In \emph{NeurIPS Datasets and Benchmarks Track}, 2024.

\bibitem[Jia et~al.(2025)Jia, You, Zhang, and Yan]{drivetransformer}
Xiaosong Jia, Junqi You, Zhiyuan Zhang, and Junchi Yan.
\newblock {DriveTransformer}: Unified transformer for scalable end-to-end
  autonomous driving.
\newblock In \emph{ICLR}, 2025.

\bibitem[Jiang et~al.(2023)Jiang, Chen, Xu, Liao, Chen, Zhou, Zhang, Liu,
  Huang, and Wang]{vad}
Bo Jiang, Shaoyu Chen, Qing Xu, Bencheng Liao, Jiajie Chen, Helong Zhou, Qian
  Zhang, Wenyu Liu, Chang Huang, and Xinggang Wang.
\newblock {VAD}: Vectorized scene representation for efficient autonomous
  driving.
\newblock In \emph{Proceedings of the IEEE/CVF International Conference on
  Computer Vision (ICCV)}, pages 8340--8350, 2023.

\bibitem[Jiang et~al.(2024)Jiang, Chen, Liao, Zhang, Yin, Zhang, Huang, Liu,
  and Wang]{senna}
Bo Jiang, Shaoyu Chen, Bencheng Liao, Xingyu Zhang, Wei Yin, Qian Zhang, Chang
  Huang, Wenyu Liu, and Xinggang Wang.
\newblock {Senna}: Bridging large vision-language models and end-to-end
  autonomous driving.
\newblock \emph{arXiv preprint arXiv:2410.22313}, 2024.

\bibitem[Jiang et~al.(2025)Jiang, Chen, Zhang, Liu, and Wang]{alphadrive}
Bo Jiang, Shaoyu Chen, Qian Zhang, Wenyu Liu, and Xinggang Wang.
\newblock {AlphaDrive}: Unleashing the power of {VLMs} in autonomous driving
  via reinforcement learning and reasoning.
\newblock \emph{arXiv preprint arXiv:2503.07608}, 2025.

\bibitem[Jin et~al.(2023)Jin, Liu, Zheng, Li, Zhao, Zhang, Zheng, Zhou, and
  Liu]{adapt}
Bu Jin, Xinyu Liu, Yupeng Zheng, Pengfei Li, Hao Zhao, Tong Zhang, Yuhang
  Zheng, Guyue Zhou, and Jingjing Liu.
\newblock {ADAPT}: Action-aware driving caption transformer.
\newblock In \emph{ICRA}, 2023.

\bibitem[Kim et~al.(2018)Kim, Rohrbach, Darrell, Canny, and Akata]{bddx}
Jinkyu Kim, Anna Rohrbach, Trevor Darrell, John Canny, and Zeynep Akata.
\newblock Textual explanations for self-driving vehicles.
\newblock In \emph{Proceedings of the European Conference on Computer Vision
  (ECCV)}, pages 563--578, 2018.

\bibitem[Lanham et~al.(2023)Lanham, Chen, Radhakrishnan, Steiner, Denison,
  Hernandez, Li, Durmus, Hubinger, Kernion, et~al.]{lanham2023faithful}
Tamera Lanham, Anna Chen, Ansh Radhakrishnan, Benoit Steiner, Carson Denison,
  Danny Hernandez, Dustin Li, Esin Durmus, Evan Hubinger, Jackson Kernion,
  et~al.
\newblock Measuring faithfulness in chain-of-thought reasoning.
\newblock \emph{arXiv preprint arXiv:2307.13702}, 2023.

\bibitem[Li et~al.(2026{\natexlab{a}})Li, Zhang, Holtz, Yu, Yang, Lai, Song,
  Geiger, and Zell]{spacedrive}
Peizheng Li, Zhenghao Zhang, David Holtz, Hang Yu, Yutong Yang, Yuzhi Lai, Rui
  Song, Andreas Geiger, and Andreas Zell.
\newblock {SpaceDrive}: Infusing spatial awareness into {VLM}-based autonomous
  driving.
\newblock In \emph{Proceedings of the IEEE/CVF Conference on Computer Vision
  and Pattern Recognition}, pages 40096--40107, 2026{\natexlab{a}}.

\bibitem[Li et~al.(2025)Li, Wang, Liu, He, Fan, and Zhang]{wote}
Yingyan Li, Yuqi Wang, Yang Liu, Jiawei He, Lue Fan, and Zhaoxiang Zhang.
\newblock End-to-end driving with online trajectory evaluation via {BEV} world
  model.
\newblock In \emph{ICCV}, 2025.

\bibitem[Li et~al.(2026{\natexlab{b}})Li, Xiong, Guo, Li, Yan, Xu, Zhou, Chen,
  Sun, Wang, Ma, Chen, Ye, Liu, and Wang]{recogdrive}
Yongkang Li, Kaixin Xiong, Xiangyu Guo, Fang Li, Sixu Yan, Gangwei Xu, Lijun
  Zhou, Long Chen, Haiyang Sun, Bing Wang, Kun Ma, Guang Chen, Hangjun Ye,
  Wenyu Liu, and Xinggang Wang.
\newblock {ReCogDrive}: A reinforced cognitive framework for end-to-end
  autonomous driving.
\newblock In \emph{ICLR}, 2026{\natexlab{b}}.
\newblock arXiv:2506.08052.

\bibitem[Luo et~al.(2026)Luo, Li, Xu, Lai, Yang, Chen, Luo, Xie, Jiang, Liu,
  Chen, Wang, and Yang]{adathinkdrive}
Yuechen Luo, Fang Li, Shaoqing Xu, Zhiyi Lai, Lei Yang, Qimao Chen, Ziang Luo,
  Zixun Xie, Shengyin Jiang, Jiaxin Liu, Long Chen, Bing Wang, and Zhi-xin
  Yang.
\newblock {AdaThinkDrive}: Adaptive thinking via reinforcement learning for
  autonomous driving.
\newblock In \emph{Proceedings of the IEEE International Conference on Robotics
  and Automation (ICRA)}, 2026.

\bibitem[Manalu et~al.(2026)Manalu, Wazir, Do, Kim, Hwang, Kidane, and
  Kim]{couplingcf}
Gabriel Manalu, Saad Wazir, Thao Do, Seongah Kim, Hyeontaek Hwang, Tolcha~Yalew
  Kidane, and Daeyoung Kim.
\newblock Measuring the reasoning-action coupling in driving
  vision-language-action models through counterfactuals.
\newblock \emph{arXiv preprint arXiv:2610.30953}, 2026.

\bibitem[Mao et~al.(2023)Mao, Qian, Ye, Zhao, and Wang]{gptdriver}
Jiageng Mao, Yuxi Qian, Junjie Ye, Hang Zhao, and Yue Wang.
\newblock {GPT-Driver}: Learning to drive with {GPT}.
\newblock \emph{arXiv preprint arXiv:2310.01415}, 2023.

\bibitem[Masson et~al.(2016)Masson, Ranchod, and
  Konidaris]{masson2016parameterized}
Warwick Masson, Pravesh Ranchod, and George Konidaris.
\newblock Reinforcement learning with parameterized actions.
\newblock In \emph{AAAI}, 2016.

\bibitem[Mayumu et~al.(2026)Mayumu, Deng, and Mukala]{vlafaithful}
Nicanor Mayumu, Xiaoheng Deng, and Patrick Mukala.
\newblock Is {VLA} reasoning faithful? {P}robing safety of chain-of-causation
  in autonomous driving models.
\newblock \emph{arXiv preprint arXiv:2605.17268}, 2026.

\bibitem[Peng et~al.(2026)Peng, Ding, You, Chen, Luo, Tian, Cao, Sharma, Xu,
  Ivanovic, Li, Wang, and Pavone]{cfvla}
Zhenghao Peng, Wenhao Ding, Yurong You, Yuxiao Chen, Wenjie Luo, Thomas Tian,
  Yulong Cao, Apoorva Sharma, Danfei Xu, Boris Ivanovic, Boyi Li, Yan Wang, and
  Marco Pavone.
\newblock Counterfactual {VLA}: Self-reflective vision-language-action model
  with adaptive reasoning.
\newblock In \emph{Proceedings of the IEEE/CVF Conference on Computer Vision
  and Pattern Recognition}, pages 4022--4031, 2026.

\bibitem[Renz et~al.(2025)Renz, Chen, Arani, and Sinavski]{simlingo}
Katrin Renz, Long Chen, Elahe Arani, and Oleg Sinavski.
\newblock {SimLingo}: Vision-only closed-loop autonomous driving with
  language-action alignment.
\newblock In \emph{CVPR}, 2025.

\bibitem[Sha et~al.(2023)Sha, Mu, Jiang, Chen, Xu, Luo, Li, Tomizuka, Zhan, and
  Ding]{languagempc}
Hao Sha, Yao Mu, Yuxuan Jiang, Li Chen, Chenfeng Xu, Ping Luo, Shengbo~Eben Li,
  Masayoshi Tomizuka, Wei Zhan, and Mingyu Ding.
\newblock {LanguageMPC}: Large language models as decision makers for
  autonomous driving.
\newblock \emph{arXiv preprint arXiv:2310.03026}, 2023.

\bibitem[Shalev-Shwartz et~al.(2017)Shalev-Shwartz, Shammah, and Shashua]{rss}
Shai Shalev-Shwartz, Shaked Shammah, and Amnon Shashua.
\newblock On a formal model of safe and scalable self-driving cars.
\newblock \emph{arXiv preprint arXiv:1708.06374}, 2017.

\bibitem[Shang et~al.(2025)Shang, Chen, Wang, Li, and Zhang]{drivedpo}
Shuyao Shang, Yuntao Chen, Yuqi Wang, Yingyan Li, and Zhaoxiang Zhang.
\newblock {DriveDPO}: Policy learning via safety {DPO} for end-to-end
  autonomous driving.
\newblock In \emph{NeurIPS}, 2025.

\bibitem[Shao et~al.(2024)Shao, Hu, Wang, Song, Waslander, Liu, and
  Li]{lmdrive}
Hao Shao, Yuxuan Hu, Letian Wang, Guanglu Song, Steven~L. Waslander, Yu Liu,
  and Hongsheng Li.
\newblock {LMDrive}: Closed-loop end-to-end driving with large language models.
\newblock In \emph{CVPR}, 2024.

\bibitem[Song et~al.(2025)Song, Jia, Liu, Pan, Zhang, Wang, Zhang, Xu, Yang,
  and Luo]{momad}
Ziying Song, Caiyan Jia, Lin Liu, Hongyu Pan, Yongchang Zhang, Junming Wang,
  Xingyu Zhang, Shaoqing Xu, Lei Yang, and Yadan Luo.
\newblock Don't shake the wheel: Momentum-aware planning in end-to-end
  autonomous driving.
\newblock In \emph{CVPR}, 2025.

\bibitem[Tancik et~al.(2020)Tancik, Srinivasan, Mildenhall, Fridovich-Keil,
  Raghavan, Singhal, Ramamoorthi, Barron, and Ng]{tancik2020fourier}
Matthew Tancik, Pratul~P. Srinivasan, Ben Mildenhall, Sara Fridovich-Keil,
  Nithin Raghavan, Utkarsh Singhal, Ravi Ramamoorthi, Jonathan~T. Barron, and
  Ren Ng.
\newblock Fourier features let networks learn high frequency functions in low
  dimensional domains.
\newblock In \emph{NeurIPS}, 2020.

\bibitem[Tian et~al.(2025)Tian, Gu, Li, Liu, Wang, Zhao, Zhan, Jia, Lang, and
  Zhao]{drivevlm}
Xiaoyu Tian, Junru Gu, Bailin Li, Yicheng Liu, Yang Wang, Zhiyong Zhao, Kun
  Zhan, Peng Jia, XianPeng Lang, and Hang Zhao.
\newblock Drivevlm: The convergence of autonomous driving and large
  vision-language models.
\newblock In \emph{Proceedings of The 8th Conference on Robot Learning}, pages
  4698--4726, 2025.

\bibitem[Turpin et~al.(2023)Turpin, Michael, Perez, and
  Bowman]{turpin2023unfaithful}
Miles Turpin, Julian Michael, Ethan Perez, and Samuel~R. Bowman.
\newblock Language models don't always say what they think: Unfaithful
  explanations in chain-of-thought prompting.
\newblock In \emph{NeurIPS}, 2023.

\bibitem[Wang et~al.(2025{\natexlab{a}})Wang, Yu, Jiang, Lan, Shi, Chang,
  Kautz, Li, and Alvarez]{omnidrive}
Shihao Wang, Zhiding Yu, Xiaohui Jiang, Shiyi Lan, Min Shi, Nadine Chang, Jan
  Kautz, Ying Li, and Jose~M. Alvarez.
\newblock {OmniDrive}: A holistic vision-language dataset for autonomous
  driving with counterfactual reasoning.
\newblock In \emph{Proceedings of the IEEE/CVF Conference on Computer Vision
  and Pattern Recognition (CVPR)}, pages 22442--22452, 2025{\natexlab{a}}.

\bibitem[Wang et~al.(2025{\natexlab{b}})Wang, Zhang, Qu, Li, Liu, and
  Huang]{diffad}
Tao Wang, Cong Zhang, Xingguang Qu, Kun Li, Weiwei Liu, and Chang Huang.
\newblock {DiffAD}: A unified diffusion modeling approach for autonomous
  driving.
\newblock \emph{arXiv preprint arXiv:2503.12170}, 2025{\natexlab{b}}.

\bibitem[Wang et~al.(2025{\natexlab{c}})]{alpamayor1}
Yan Wang et~al.
\newblock {Alpamayo-R1}: Bridging reasoning and action prediction for
  generalizable autonomous driving in the long tail.
\newblock \emph{arXiv preprint arXiv:2511.00088}, 2025{\natexlab{c}}.

\bibitem[Wu et~al.(2022)Wu, Jia, Chen, Yan, Li, and Qiao]{tcp}
Penghao Wu, Xiaosong Jia, Li Chen, Junchi Yan, Hongyang Li, and Yu Qiao.
\newblock Trajectory-guided control prediction for end-to-end autonomous
  driving: A simple yet strong baseline.
\newblock In \emph{NeurIPS}, 2022.

\bibitem[Wu et~al.(2025)Wu, Li, Hermans, Ramos, Bajcsy, and
  P{\'e}rez-D'Arpino]{seal}
Yilin Wu, Anqi Li, Tucker Hermans, Fabio Ramos, Andrea Bajcsy, and Claudia
  P{\'e}rez-D'Arpino.
\newblock Do what you say: Steering vision-language-action models via runtime
  reasoning-action alignment verification.
\newblock \emph{arXiv preprint arXiv:2510.16281}, 2025.

\bibitem[Xie et~al.(2025)Xie, Kong, Dong, Sima, Zhang, Chen, Liu, and
  Pan]{drivebench}
Shaoyuan Xie, Lingdong Kong, Yuhao Dong, Chonghao Sima, Wenwei Zhang, Qi~Alfred
  Chen, Ziwei Liu, and Liang Pan.
\newblock Are {VLMs} ready for autonomous driving? an empirical study from the
  reliability, data, and metric perspectives.
\newblock In \emph{ICCV}, 2025.

\bibitem[Xiong et~al.(2026)Xiong, Ye, Yaman, Cheng, Lu, Luo, Jacobs, and
  Ren]{unidrivewm}
Zhexiao Xiong, Xin Ye, Burhan Yaman, Sheng Cheng, Yiren Lu, Jingru Luo, Nathan
  Jacobs, and Liu Ren.
\newblock {UniDrive-WM}: Unified understanding, planning and generation world
  model for autonomous driving.
\newblock \emph{arXiv preprint arXiv:2601.04453}, 2026.

\bibitem[Xu et~al.(2024)Xu, Zhang, Xie, Zhao, Guo, Wong, Li, and
  Zhao]{drivegpt4}
Zhenhua Xu, Yujia Zhang, Enze Xie, Zhen Zhao, Yong Guo, Kwan-Yee~K. Wong,
  Zhenguo Li, and Hengshuang Zhao.
\newblock {DriveGPT4}: Interpretable end-to-end autonomous driving via large
  language model.
\newblock \emph{IEEE Robotics and Automation Letters}, 2024.

\bibitem[Yang et~al.(2024)Yang, Yang, Hui, Zheng, Yu, Zhou, Li, Li, Liu, Huang,
  et~al.]{qwen2}
An Yang, Baosong Yang, Binyuan Hui, Bo Zheng, Bowen Yu, Chang Zhou, Chengpeng
  Li, Chengyuan Li, Dayiheng Liu, Fei Huang, et~al.
\newblock {Qwen2} technical report.
\newblock \emph{arXiv preprint arXiv:2407.10671}, 2024.

\bibitem[Yang et~al.(2025{\natexlab{a}})Yang, Chai, Jia, Li, Shao, Zhu, Su, and
  Yan]{drivemoe}
Zhenjie Yang, Yilin Chai, Xiaosong Jia, Qifeng Li, Yuqian Shao, Xuekai Zhu,
  Haisheng Su, and Junchi Yan.
\newblock {DriveMoE}: Mixture-of-experts for vision-language-action model in
  end-to-end autonomous driving.
\newblock \emph{arXiv preprint arXiv:2505.16278}, 2025{\natexlab{a}}.

\bibitem[Yang et~al.(2025{\natexlab{b}})Yang, Jia, Li, Yang, Yao, and
  Yan]{raw2drive}
Zhenjie Yang, Xiaosong Jia, Qifeng Li, Xue Yang, Maoqing Yao, and Junchi Yan.
\newblock {Raw2Drive}: Reinforcement learning with aligned world models for
  end-to-end autonomous driving (in {CARLA} v2).
\newblock In \emph{NeurIPS}, 2025{\natexlab{b}}.

\bibitem[Yu et~al.(2026)Yu, Feng, and Elbaum]{drivethoughts}
Tian Yu, Lu Feng, and Sebastian Elbaum.
\newblock Drive the thoughts: Runtime monitoring of {VLA} reasoning-trajectory
  consistency.
\newblock \emph{arXiv preprint arXiv:2608.29583}, 2026.

\bibitem[Zhang et~al.(2026)Zhang, Yuan, Wu, Chang, Cai, Zeng, Shi, Wang, Zhang,
  and Xu]{minddriver}
Lingjun Zhang, Yujian Yuan, Changjie Wu, Xinyuan Chang, Xin Cai, Shuang Zeng,
  Linzhe Shi, Sijin Wang, Hang Zhang, and Mu Xu.
\newblock {MindDriver}: Introducing progressive multimodal reasoning for
  autonomous driving.
\newblock In \emph{Proceedings of the IEEE/CVF Conference on Computer Vision
  and Pattern Recognition (CVPR)}, 2026.

\bibitem[Zheng et~al.(2024)Zheng, Song, Guo, Zhang, and Chen]{genad}
Wenzhao Zheng, Ruiqi Song, Xianda Guo, Chenming Zhang, and Long Chen.
\newblock {GenAD}: Generative end-to-end autonomous driving.
\newblock In \emph{ECCV}, pages 87--104, 2024.

\bibitem[Zheng et~al.(2026)Zheng, Huang, Sun, Li, and Zhao]{drivema}
Weicheng Zheng, Yixin Huang, Qiao Sun, Derun Li, and Hang Zhao.
\newblock {DriveMA}: Driving vision-language-action models with verifiable
  meta-actions.
\newblock \emph{arXiv preprint arXiv:2605.31271}, 2026.

\bibitem[Zhou et~al.(2026)Zhou, Han, Yang, Ma, Tresp, and Knoll]{opendrivevla}
Xingcheng Zhou, Xuyuan Han, Feng Yang, Yunpu Ma, Volker Tresp, and Alois Knoll.
\newblock Opendrivevla: Towards end-to-end autonomous driving with large vision
  language action model.
\newblock In \emph{Proceedings of the AAAI Conference on Artificial
  Intelligence (AAAI)}, pages 13782--13790, 2026.

\bibitem[Zhou et~al.(2025)Zhou, Cai, Zhao, Zhang, Huang, Zhou, et~al.]{autovla}
Zewei Zhou, Tianhui Cai, Seth~Z. Zhao, Yun Zhang, Zhiyu Huang, Bolei Zhou,
  et~al.
\newblock {AutoVLA}: A vision-language-action model for end-to-end autonomous
  driving with adaptive reasoning and reinforcement fine-tuning.
\newblock In \emph{Advances in Neural Information Processing Systems
  (NeurIPS)}, 2025.
\newblock arXiv:2506.13757.

\end{thebibliography}
}

\clearpage
\setcounter{section}{0}
\setcounter{table}{0}
\setcounter{figure}{0}
\setcounter{equation}{0}
\setcounter{proposition}{0}
\suppnumbering
\def\theHsection{supp.\arabic{section}}
\def\theHtable{supp.\arabic{table}}
\def\theHfigure{supp.\arabic{figure}}
\def\theHequation{supp.\arabic{equation}}
\def\theHproposition{supp.\arabic{proposition}}
\makeatletter
\setlength{\@fptop}{0pt}
\setlength{\@dblfptop}{0pt}
\makeatother
\hyphenation{MindDrive SimLingo DriveTransformer}

\maketitlesupplementary
\section{Overview}
\label{supp:sec:overview}

We provide supplementary material to complement the main paper. It is organized as follows:
\begin{itemize}
\item \cref{supp:sec:questions}: Potential questions raised by the main paper.
\item \cref{supp:sec:data}: Details on the counterfactual decision-action data.
\item \cref{supp:sec:impl}: Implementation and training details.
\item \cref{supp:sec:results}: More results.
\end{itemize}
The tables, figures, equations and algorithms of this document are numbered with the prefix A, and its sections with letters.
All other numbers refer to the main paper.

\section{Questions}
\label{supp:sec:questions}

\question{What are the differences between \method and MindDrive?}
\label{supp:q:minddrive}

\method adopts the architecture of MindDrive~\cite{minddrive}, in which a Decision Expert and an Action Expert share one VLM, and initializes its Action Expert from the released MindDrive checkpoint (\cref{supp:sec:training}).
In MindDrive, the decision is a speed token that selects one of several candidate trajectories (\cref{eq:select}).
The token is an index, and no term requires the selected candidate to follow it.
The candidates of MindDrive are nearly identical on most frames, and its output follows a forced token on a minority of the frames (\cref{supp:sec:intervention}).
In \method, the decision is a commitment $(d,\theta)$ that the deterministic rule $\Phi$ reads from any trajectory.
The \codebook maps the commitment to a code with a kinematic meaning, and the consistency losses supervise every admissible decision of a scene with a trajectory that realizes it.
The decisions of a scene then yield distinct trajectories, and the trajectory follows a forced decision on nearly all frames (\cref{supp:tab:follow}).
The supervision also differs: the decision labels of \method are read from the future motion of the expert, so the label and the trajectory agree by construction, and the records are reviewed (\cref{supp:sec:data}).

\question{Is the consistency between the decision and the trajectory guaranteed at inference?}
\label{supp:q:guarantee}

It is learned and not guaranteed.
We relax the indicator of \cref{eq:hard} into the consistency losses of \cref{eq:cons} and apply no projection or rejection at inference.
We therefore measure the consistency in two ways: with the say-do gap in closed loop and in open loop (\cref{tab:gap} and \cref{supp:sec:gapfull}) and with interventions that force the decision to another value (\cref{supp:sec:intervention}).
In both, the trajectory departs from the decision on few frames.
Since $\Phi$ is a deterministic rule on the planned trajectory, a remaining violation can be detected at run time.
We do not use such a check in any reported result.

\question{Why is conditioning the Action Expert on the decision not sufficient?}
\label{supp:q:conditioning}

On expert data, every scene comes with one decision, and this decision is a function of the scene.
A planner can then reproduce the expert from its hidden state $h_a$ alone, and the imitation loss does not constrain its output under any other decision.
We state this formally in \cref{supp:sec:shortcut}.
This agrees with what we observe for MindDrive, whose output follows a forced speed token on a minority of the frames (\cref{fig:teaser}b).
The counterfactual executions add every admissible decision of a scene to the training data, each with its own target, so only the decision tells the targets apart.
Without them, the say-do gap rises (\cref{tab:ablation}), the trajectories of different decisions move closer together, and the trajectory follows a forced decision far less often (\cref{supp:tab:ablation}).

\question{How does the counterfactual data differ from the Action Dreaming of SimLingo?}
\label{supp:q:simlingo}

SimLingo~\cite{simlingo} simulates alternative futures to train the planner to follow external language instructions.
These futures are paired with instructions, and not with the decision that the model itself states in its commentary.
In our interventions on its released checkpoint, the trajectory follows an explicit instruction on nearly all frames, but it follows the commentary far less often when the commentary is forced to another decision (\cref{supp:tab:probes}).
In \method, the counterfactual executions are paired with the decision of the model itself.
The same records supervise the Decision Expert, through the admissible set, and the Action Expert, through the counterfactual executions.
The admissible set is computed from the logged scene, so the alternatives are limited to the decisions that pass its collision check and its traffic rules (\cref{supp:sec:pipeline}).

\question{Why does the say-exec gap remain large when the say-do gap is small?}
\label{supp:q:sayexec}

The two gaps compare the stated decision with different trajectories: the say-do gap with the trajectory that is planned at the same step, which the consistency constraint covers, and the say-exec gap with the trajectory that the vehicle executes afterward.
Since the model re-plans at every step, the executed trajectory results from many consecutive plans, and a decision that is stated at one step can be replaced at the next.
The planned and the executed trajectory also differ for the baselines, as their do-exec gap shows (\cref{supp:tab:strata}).
\method lowers the say-exec gap, but a large part of it remains (\cref{tab:gap}).
This remainder concerns the consistency of the decisions of consecutive steps, which the per-step constraint of \cref{eq:hard} does not cover, and we leave it to future work.

\question{Why does the decision space contain only speed decisions?}
\label{supp:q:speed}

The Action Expert has a path branch and a speed branch (\cref{fig:framework}).
In the annotations of MindDrive~\cite{minddrive}, the model on which \method is built, the path token is the navigation command of Bench2Drive, so the decision that the model makes by itself is the speed decision.
The released checkpoint confirms this: its path token equals the navigation command on the validation frames, and its open-loop error comes almost entirely from the speed along the path (\cref{supp:sec:model}).
We therefore keep the path branch frozen and constrain the speed decision.
The formulation does not depend on this choice: any decision that a deterministic rule can read from a trajectory can be stated as a commitment.
We leave path decisions to future work.

\question{What is the role of the VLM in the data pipeline?}
\label{supp:q:vlm}

The VLM does not produce the labels.
Rule-based functions generate all four records of a frame from the raw log. The decision label is read from the future motion of the expert, and the counterfactual executions, the admissible set and the perception facts are computed from the logged scene (\cref{supp:sec:pipeline}).
The VLM~\cite{qwen3vl} then acts in two roles, as a blind reviewer that judges the records without knowing their origin or the verdict of the rule-based functions, and as an editor that revises reasoning text.
When the reviewer disagrees because of an error that originates in a rule or a template, we correct it at its source and regenerate the records, so the correction applies to every frame.
In the example of \cref{fig:pipeline}, the reviewer judges that keeping the current speed toward a red light is illegal, and the rule that builds the admissible set is corrected to forbid \textsc{keep} when the stop line is within reach (\cref{supp:eq:reach}).
\cref{supp:fig:records} shows a record of this kind before and after the correction.

\section{Details on the Counterfactual Decision-Action Data}
\label{supp:sec:data}\suppanchor{\SuppDataRef}{\thesection}

The language annotations on which MindDrive~\cite{minddrive} is trained are not checked against the expert trajectory.
Their speed token agrees with the decision that the expert realizes over the next 3\,s on 40.0\% of the labeled frames (mapping of \cref{supp:sec:mapping}).
To supply the supervision that the consistent language-action mapping needs, we build an automatic data pipeline on the raw logs of the Bench2Drive base set~\cite{bench2drive}, without new recordings.
This section details the pipeline (\cref{supp:sec:pipeline}), the resulting dataset (\cref{supp:sec:stats}) and the mapping that expresses the decisions of existing models in our decision space (\cref{supp:sec:mapping}).

\subsection{Data Pipeline}
\label{supp:sec:pipeline}
\begin{algorithm}[t]
\caption{\textbf{Counterfactual decision-action data pipeline.}}
\label{supp:alg:pipeline}
\begin{algorithmic}[1]
\Require raw logs of the Bench2Drive base set
\Ensure records $(\bar d,\bar\theta)$, $\{\bar a_k\}$, $\Dadm$, $F$ of every labeled frame
\Phase{Step 1: generate, by rule-based functions}
\ForAll{frames}
  \State $a^{\mathrm{e}}\gets$ next 3\,s of the expert trajectory
  \State $(\bar d,\bar\theta)\gets\big(\Phi(a^{\mathrm{e}}),\Theta(a^{\mathrm{e}})\big)$; \ $\bar a_{\bar d}\gets a^{\mathrm{e}}$
  \ForAll{$k\in\Dset\setminus\{\bar d\}$}
    \State $\bar a_k\gets$ path of $a^{\mathrm{e}}$, re-timed with a speed profile of decision $k$
    \State roll out $\bar a_k$ on the occupancy grid; infeasible if it collides
  \EndFor
  \State $\Dadm\gets\{k:\bar a_k$ feasible, no traffic rule forbids $k\}$
  \State $F\gets$ fact sentences rendered from the logged scene
\EndFor
\Phase{Step 2: review, by the VLM as a blind reviewer}
\ForAll{records of the review sample}
  \State judge the decision sentence, the fact sentences and the alternative decisions; flag the record on a contradiction
\EndFor
\Phase{Step 3: refine}
\ForAll{flagged records}
  \If{the error originates in a rule or a template}
    \State correct it at its source, regenerate and review again
  \Else
    \State revise the reasoning text, by the VLM as an editor
  \EndIf
\EndFor
\State human experts check a random sample of the records
\end{algorithmic}
\end{algorithm}

\begin{table*}[t]
\centering
\caption{\textbf{Templates of the decision sentence and of the perception facts in the prompt of the Decision Expert.} \slot{$\cdot$} is a slot that is filled from the record for training and, at inference, from the vehicle state and the outputs of the model, and $\mid$ separates alternatives.}
\label{supp:tab:templates}
\begin{promptbox}
\prompthead{Type 1: Question and decision sentence}
Question: ``What actions should the car be taking?''\\
Decision sentence: ``The car should be \slot{decision token} \slot{value token} \slot{path token}'', for example ``The car should be \texttt{<d\_slow>} \texttt{<v=4>} \texttt{<lanefollow>}''.\\
Decision tokens: \texttt{<d\_stop>}, \texttt{<d\_slow>}, \texttt{<d\_keep>}, \texttt{<d\_accel>}.\\
Value tokens: \texttt{<s=$i$>} for a stopping distance of $2i$\,m, $i=0,\dots,40$ (\textsc{stop}); \texttt{<v=$i$>} for an end speed of $i$\,m/s, $i=0,\dots,15$ (\textsc{slow}, \textsc{keep}, \textsc{accel}).\\
Path tokens: the six path tokens of MindDrive, selected by the navigation command at inference.
\promptsep
\prompthead{Type 2: Ego state}
``Ego speed: \slot{$v_0$}\,m/s. Navigation: \slot{turn left $\mid$ turn right $\mid$ go straight $\mid$ follow the lane $\mid$ change lanes to the left $\mid$ change lanes to the right}.''
\promptsep
\prompthead{Type 3: Perception facts, in this order}
1. Traffic light: ``Traffic light: \slot{red $\mid$ yellow $\mid$ green}, affects ego.'' $\mid$ ``Traffic light: \slot{state}, not affecting ego.'' $\mid$ ``Traffic light: none.''\\
2. Objects: ``Objects: \slot{object}; \dots; \slot{object}.'' $\mid$ ``Objects: none.'', where an object reads ``\slot{class} \slot{$n$\,m ahead $\mid$ alongside $\mid$ $n$\,m behind}, \slot{in lane $\mid$ $n$\,m left $\mid$ $n$\,m right}, \slot{stationary $\mid$ moving ahead at $v$\,m/s $\mid$ approaching at $v$\,m/s $\mid$ crossing left at $v$\,m/s $\mid$ crossing right at $v$\,m/s}''. The list has up to six objects: the nearest cars, vans, trucks, bicycles, pedestrians, traffic cones and traffic signs within 40\,m that are at most 12\,m to the side and 3\,m behind.\\
3. Lead vehicle: ``Lead vehicle: \slot{class} \slot{$n$}\,m ahead, \slot{closing at $v$\,m/s, time to collision $t$\,s $\mid$ pulling away at $v$\,m/s $\mid$ same speed as ego}, time gap \slot{$t$}\,s.'' $\mid$ ``Lead vehicle: none within 40\,m.''\\
4. Vulnerable road user: ``Vulnerable road user in the corridor: \slot{class} \slot{$n$}\,m ahead, \slot{in lane $\mid$ $n$\,m left $\mid$ $n$\,m right}, \slot{stationary $\mid$ crossing left at $v$\,m/s $\mid$ crossing right at $v$\,m/s $\mid$ moving along the road at $v$\,m/s}.'' $\mid$ ``Vulnerable road user in the corridor: none within 30\,m.''
\end{promptbox}
\end{table*}

As shown in \cref{fig:pipeline}, the pipeline consists of three steps, followed by a check of a random sample by human experts: rule-based functions generate the records of every frame, a VLM reviews them, and flagged records are refined (\cref{supp:alg:pipeline}).

\paragraph{Decision label.}
Let $a=(p_1,\dots,p_{N_s})$ be a trajectory over the next 3\,s, whose positions follow the current position $p_0$ at intervals of $\Delta t=0.5$\,s.
Its speed profile $w(a)=(v_1,\dots,v_{N_s})$ consists of the step speeds $v_j=\|p_j-p_{j-1}\|/\Delta t$, and $v_0$ is the current speed.
The rule $\Phi$ reads the decision that the trajectory $a$ realizes from its end speed $v_{N_s}$, the mean speed over the last 0.5\,s:
\begin{equation}
\label{supp:eq:phi}
\Phi(a)=
\begin{cases}
\textsc{stop}, & v_{N_s}<\epsilon_v,\\
\textsc{slow}, & v_{N_s}-v_0<-\delta_v,\\
\textsc{accel}, & v_{N_s}-v_0>\delta_v,\\
\textsc{keep}, & \text{otherwise},
\end{cases}
\end{equation}
where the cases are evaluated from top to bottom, with $\epsilon_v=0.5$\,m/s and $\delta_v=1$\,m/s.
The target value of the decision is read from the same trajectory:
\begin{equation}
\label{supp:eq:theta}
\Theta(a)=
\begin{cases}
\sum_{j=1}^{N_s}\|p_j-p_{j-1}\|, & \Phi(a)=\textsc{stop},\\
v_{N_s}, & \text{otherwise}.
\end{cases}
\end{equation}
For \textsc{stop}, it is the distance that the vehicle travels within the 3\,s.
The vehicle is at rest at the end of this horizon, so the distance is its stopping distance.
The label of a frame is the commitment $(\bar d,\bar\theta)=(\Phi(a^{\mathrm{e}}),\Theta(a^{\mathrm{e}}))$ of the expert trajectory $a^{\mathrm{e}}$.
It is rendered as the decision sentence by a template (\cref{supp:tab:templates}), for example ``The car should be \texttt{<d\_stop>} \texttt{<s=8>} \texttt{<lanefollow>}'' for a stop within 16\,m.
The figures show this sentence in plain words, such as ``stop within 16\,m''.
The label and the trajectory agree by construction.
A frame carries a label if the 1\,s before it and the 3\,s after it lie within its clip.

\paragraph{Counterfactual executions.}
For every other decision $k\neq\bar d$, the path of the expert is kept and re-timed with a speed profile of decision $k$:
\begin{equation}
\label{supp:eq:profiles}
v_j=
\begin{cases}
v_0, & \textsc{keep},\\
v_0-\tfrac{j}{N_s}\max(1.5,\,0.3\,v_0), & \textsc{slow},\\
v_0+\tfrac{j}{N_s}\,1.5, & \textsc{accel},
\end{cases}
\end{equation}
with all speeds in m/s.
For \textsc{stop}, the profile has a constant deceleration of $\max\big(2.5\,\mathrm{m/s^2},\,v_0/(2.475\,\mathrm{s})\big)$, which brings the vehicle to rest before the last step.
The positions follow from moving along the polyline of the expert trajectory by the traveled distance, and beyond its end the path continues straight in its last direction.
This gives a trajectory $\bar a_k$ with $\Phi(\bar a_k)=k$ and the target value $\theta_k=\Theta(\bar a_k)$.
Its end speed keeps a margin of at least 0.3\,m/s from the thresholds of $\Phi$.
Only the speed along the path changes, so the executions of a frame differ in exactly the quantity that the decision describes.
A decision has no execution if it cannot be realized from the current speed: \textsc{keep} below 0.8\,m/s, \textsc{slow} below 2.3\,m/s, \textsc{accel} above 11.5\,m/s, and \textsc{stop} if it requires a deceleration above 4.45\,m/s$^2$.

Each $\bar a_k$ is rolled out against the logged motion of the other agents on the occupancy grid of the open-loop planning metric~\cite{stp3}, which has $200\times200$ cells of 0.5\,m.
The boxes of the vehicles, cyclists and pedestrians are drawn at their logged positions at each of the six future steps, and the footprint of the ego vehicle, 4.08\,m by 1.85\,m, is placed at the waypoint of the same step.
The execution $\bar a_k$ is infeasible if the footprint overlaps a box at any step.
A collision with an agent that follows the ego vehicle, with its center more than 1\,m behind and within 2\,m to the side, does not count, since a logged follower cannot react to a slower ego vehicle.
If the \textsc{stop} execution collides, decelerations of 3.0, 3.5 and 4.45\,m/s$^2$ are tried in turn.
The drivable area is not checked.

\paragraph{Admissible set.}
The admissible set keeps the decisions whose execution is feasible and that no traffic rule forbids:
\begin{equation}
\label{supp:eq:adm}
\Dadm=\big\{k\in\Dset:\ \bar a_k\ \text{is feasible},\ k\ \text{is legal}\big\},
\end{equation}
where $\bar a_{\bar d}=a^{\mathrm{e}}$ is the expert trajectory, and a decision is legal if none of three traffic rules forbids it.
The first two concern a red light that governs the ego lane: \textsc{accel} is forbidden at such a light, and \textsc{keep} as well when its stop line is within reach.
The third forbids \textsc{keep} and \textsc{accel} when a stop sign of the ego lane is within reach and the vehicle has not stopped for it yet.
Within reach of the stop line of a red light or of a stop sign, both \textsc{keep} and \textsc{accel} are forbidden, as stated in \cref{sec:method:data}.
A stop line or a stop sign at a distance $y$ ahead is within reach of a vehicle that moves at $v_0\geq1$\,m/s if
\begin{equation}
\label{supp:eq:reach}
0<y\leq v_0\cdot3\,\mathrm{s}+5\,\mathrm{m},
\end{equation}
the distance that \textsc{keep} covers within the horizon plus a margin.
\textsc{stop} and \textsc{slow} are never forbidden.
The state of the light and the distances to the stop line and to the stop sign are taken from the annotations of the simulator.
If a rule forbids the decision of the expert, this decision is not in $\Dadm$, and the label of the Decision Expert is corrected (\cref{supp:sec:training}).
If no decision is admissible, $\Dadm$ falls back to $\{\bar d\}$, which occurs on seven training frames.

\paragraph{Perception facts.}
The perception facts $F$ are short sentences, such as ``Traffic light: red, affects ego.'' and ``Objects: car 12\,m ahead, in lane, stationary.''
For training, they are rendered from the annotated boxes and the annotated state of the traffic light.
At inference, the same function renders them from the outputs of the detection heads of the model: the boxes with a score above 0.3 and the detected traffic light that governs the ego lane (\cref{supp:tab:templates}).

\begin{table*}[t]
\centering
\caption{\textbf{Prompts fed into Qwen3-VL to review the records.} The six camera images precede the text. \slot{$\cdot$} is a slot that is filled from the record. For Prompt 1, the list of facts ends with the rendering conventions of the sentence under review, such as the rounding of its numbers and the thresholds of the decisions.}
\label{supp:tab:prompts}
\begin{promptbox}
\prompthead{Prompt 1: Review of the decision sentence and of the fact sentences}
\textit{System:} You audit answers written for a driving dataset. You check every factual claim in the answer against six camera images of the scene and a list of ground-truth facts from the vehicle's sensors and annotations. The ground-truth facts are reliable: when they disagree with your visual impression, trust the facts. Claims that neither the images nor the facts can settle are not errors. Be strict but fair, and answer only with JSON.\\[2pt]
\textit{User:} Question that was asked: \slot{question}\\
Answer to audit: \slot{sentence}\\
Ground-truth facts: \slot{facts}\\
Tasks:\\
1. List each claim in the answer that CONTRADICTS the images or the ground-truth facts. Types: ``existence'' (says an object is present that is not there, or denies an object that is there), ``count'', ``direction'' (wrong position or side), ``safety'' (calls the situation safe/clear although a listed hazard exists, or the reverse), ``ego\_action'' (the ego vehicle's speed or turning behaviour disagrees with its actual next 3 s), ``traffic\_light'', ``invisible\_mention'' (names an object that cannot be seen in any image), ``other''. Do not list omissions here.\\
2. ``omits\_critical'': true only if the answer, given what the question asks, leaves out a listed hazard or a pedestrian or vehicle in the ego vehicle's path that a careful driver would mention.\\
3. ``fluency'': 1 (broken or incoherent) to 5 (natural and coherent).\\
Reply with JSON only: \{``contradictions'': [\{``claim'': ``...'', ``type'': ``...'', ``why'': ``...''\}], ``omits\_critical'': false, ``omission'': ``'', ``fluency'': 5\}\\[2pt]
\textit{Question for the decision sentence:} What should the car do over the next 3 seconds?\\
\textit{Question for the fact sentences:} Describe the traffic light state and the relevant objects ahead of the ego vehicle.
\promptsep
\prompthead{Prompt 2: Review of the admissible set}
\textit{System:} You assess candidate driving actions for an automated vehicle. You see six camera images of the scene and a list of ground-truth facts from the vehicle's sensors and annotations. The ground-truth facts are reliable: when they disagree with your visual impression, trust the facts. An action is admissible when carrying it out over the next 3 seconds obeys the traffic rules, in particular the traffic light controlling the ego lane, and does not lead to a collision with another road user. An action can be admissible even if it is not the most efficient choice. Be strict but fair, and answer only with JSON.\\[2pt]
\textit{User:} Ground-truth facts: \slot{facts}\\
The candidate action keeps the ego vehicle on its current path and only changes its speed.\\
Over the next 3 seconds, would it be safe and legal for the ego vehicle to \slot{action} while \slot{navigation}? Consider the traffic light, other road users and the ego speed (\slot{$v_0$} m/s). Answer JSON only: \{``admissible'': true/false, ``reason'': ``...''\}
\end{promptbox}
\end{table*}

\paragraph{Review.}
The contradictions of an annotation are also learned by the model, so the records are reviewed.
The reviewer is the 32B version of Qwen3-VL~\cite{qwen3vl} with greedy decoding.
It receives the six camera images of a frame, a list of ground-truth facts from the simulator and the prompts of \cref{supp:tab:prompts}, and it answers in JSON.
The review covers a sample of the validation records: the decision sentence and the fact sentences of 600 frames, 150 for each decision, and the admissibility of 300 pairs of a frame and a decision, of which 150 are admissible, 75 are forbidden at a red light and 75 are infeasible.
The review is blind: the items are shuffled and carry random identifiers, and the reviewer is told neither the origin of a sentence nor the verdict of the rule-based functions.
For the admissibility items, the future motion of the ego vehicle is withheld as well.
The reviewer reads the decision in plain words.
As a reference, the decision of the original annotation of each frame is rendered and reviewed in the same way.
The reviewer finds a contradiction between the decision sentence and the motion of the ego vehicle in 62.5\% of the original annotations and in 26.7\% of ours.
It agrees with the admissible set on 66.7\% of the pairs: it rejects all 75 decisions that a red light forbids and 62.7\% of the infeasible ones, and it accepts 52.0\% of the admissible ones.

\paragraph{Refine.}
A disagreement of the reviewer is traced to its origin, since the reviewer can err as well.
An error that originates in a rule or a template is corrected at its source, the records of all frames are regenerated, and the sample is reviewed again (dashed arrow in \cref{fig:pipeline}).
Two of the three traffic rules above entered the pipeline in this way.
The first version of the rules forbade only \textsc{accel} at a red light.
The reviewer rejected \textsc{keep} when a red light was a short distance ahead, and it rejected decisions that take the vehicle past a stop sign.
The rule on the reach of a red light and the stop-sign rule were added, which changed the admissible set of 3.4\% of the labeled training frames and of 2.7\% of the labeled validation frames.
The sentence templates were revised in the same way over four rounds of review.
Reasoning text that contradicts itself is revised by the VLM, which then acts as an editor.
Finally, human experts check a random sample of the records.

\subsection{Dataset Statistics}
\label{supp:sec:stats}
\begin{table}[t]
\centering
\caption{\textbf{Statistics of the dataset on the Bench2Drive base set.} Shares refer to the labeled frames. The decision labels are those read from the expert trajectory. An execution can be both infeasible and forbidden by a traffic rule.}
\label{supp:tab:stats}
\small
\setlength{\tabcolsep}{4pt}
\resizebox{\linewidth}{!}{%
\begin{tabular}{@{}lrr@{}}
\toprule
 & Training & Validation \\
\midrule
Clips & 950 & 50 \\
Frames & 234{,}769 & 12{,}806 \\
Frames with a label & 196{,}769 & 10{,}806 \\
\midrule
\multicolumn{3}{@{}l}{\emph{Decision labels (\%)}} \\
\quad \textsc{stop} & 37.5 & 43.8 \\
\quad \textsc{slow} & 12.7 & 10.9 \\
\quad \textsc{keep} & 20.7 & 18.0 \\
\quad \textsc{accel} & 29.1 & 27.4 \\
\midrule
\multicolumn{3}{@{}l}{\emph{Counterfactual executions}} \\
\quad Generated & 422{,}141 & 22{,}004 \\
\quad Infeasible (colliding) & 48{,}730 & 2{,}113 \\
\quad Forbidden by a traffic rule & 40{,}991 & 2{,}646 \\
\quad Admissible & 334{,}315 & 17{,}324 \\
\midrule
\multicolumn{3}{@{}l}{\emph{Admissible set}} \\
\quad Frames with 1 decision (\%) & 22.5 & 25.5 \\
\quad Frames with 2 decisions (\%) & 27.9 & 29.2 \\
\quad Frames with 3 decisions (\%) & 9.0 & 7.4 \\
\quad Frames with 4 decisions (\%) & 40.5 & 37.9 \\
\quad Mean number of decisions & 2.67 & 2.58 \\
\quad Red light governs the ego lane (\%) & 18.3 & 22.8 \\
\quad Stop line or stop sign within reach (frames) & 7{,}107 & 324 \\
\quad Decision of the expert forbidden (\%) & 2.4 & 2.6 \\
\midrule
\multicolumn{3}{@{}l}{\emph{Review by the VLM}} \\
\quad Frames with reviewed sentences & -- & 600 \\
\quad Reviewed pairs of a frame and a decision & -- & 300 \\
\bottomrule
\end{tabular}}
\end{table}

\begin{figure*}[t]
\centering
\includegraphics[width=\linewidth]{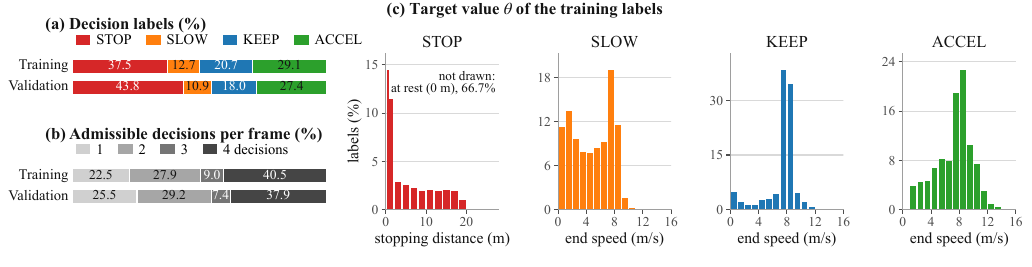}
\caption{\textbf{Statistics of the counterfactual decision-action data.} (a) Share of the four decisions among the labels of the training and of the validation frames. (b) Share of the labeled frames with one to four admissible decisions. (c) Target value of the training labels in bins of 2\,m and of 1\,m/s, as a share of the labels of each decision: the stopping distance of \textsc{stop} and the end speed of \textsc{slow}, \textsc{keep} and \textsc{accel}. For \textsc{stop}, the frames on which the vehicle is at rest, 66.7\% of the \textsc{stop} labels, are not drawn.}
\label{supp:fig:stats}
\end{figure*}

The resulting dataset covers the base set of Bench2Drive, with 950 clips and 234{,}769 frames for training and 50 clips and 12{,}806 frames for open-loop validation (\cref{supp:tab:stats}).
A label exists for 196{,}769 training frames and 10{,}806 validation frames.
Every labeled frame carries four records: the decision label, the counterfactual executions of the other decisions, the admissible set and the perception facts.
\textsc{stop} is the most frequent decision, with 37.5\% of the training labels, because the vehicle is at rest in many frames.
\textsc{accel} follows with 29.1\%, \textsc{keep} with 20.7\% and \textsc{slow} with 12.7\%.
For the training frames, 422{,}141 counterfactual executions are generated, of which 334{,}315 are admissible.
The labeled training frames have 2.67 admissible decisions on average, and 77.5\% of them have at least two, so that the Action Expert sees most scenes under more than one decision.
A red light governs the ego lane on 18.3\% of the labeled training frames, and a traffic rule forbids the decision of the expert on 2.4\%.
\cref{supp:fig:stats} shows the distribution of the decisions and of their target values.

\subsection{Mapping the Decisions of Existing Models}
\label{supp:sec:mapping}
\begin{table*}[t]
\centering
\caption{\textbf{Mapping of the stated decisions of existing models to the decision space $\Dset$.} Yield: the yield decision to which each decision belongs. For SimLingo, the number of distinct action phrases of each decision is given with examples. ORION states only the yield decision, which is read as described in \cref{supp:sec:mapping}.}
\label{supp:tab:mapping}
\small
\setlength{\tabcolsep}{5pt}
\begin{tabular}{@{}ll>{\raggedright\arraybackslash}p{5.0cm}>{\raggedright\arraybackslash}p{7.8cm}@{}}
\toprule
Decision & Yield & MindDrive speed tokens & SimLingo commentary phrases \\
\midrule
\textsc{stop} & yield & \texttt{stop} & \prov{57}, e.g.\ ``stop now'', ``remain stopped'', ``wait for a gap in the traffic'' \\
\textsc{slow} & yield & \texttt{slow\_down}, \texttt{slow\_down\_rapidly} & \prov{21}, e.g.\ ``decelerate'', ``brake gently'', ``lower your speed'' \\
\textsc{keep} & no yield & \texttt{maintain\_slow\_speed}, \texttt{maintain\_moderate\_speed}, \texttt{maintain\_fast\_speed} & \prov{61}, e.g.\ ``continue at the same speed'', ``do not accelerate'' \\
\textsc{accel} & no yield & \texttt{speed\_up} & \prov{21}, e.g.\ ``accelerate'', ``drive faster'', ``gain speed'' \\
\bottomrule
\end{tabular}
\end{table*}

To measure the say-do gap of an existing model (\cref{tab:gap}), its stated decision has to be expressed in the decision space $\Dset$.
\cref{supp:tab:mapping} lists the mapping.

\paragraph{Yield decision.}
The yield decision, the yielding action of \cref{tab:gap}, groups \textsc{stop} and \textsc{slow} as yield, and \textsc{keep} and \textsc{accel} as no yield.
It is the only decision that all evaluated models state, and \cref{tab:gap} reports it next to the speed decision.

\paragraph{MindDrive.}
MindDrive~\cite{minddrive} states its decision as one of seven speed tokens.
Four of them name a stop or a change of speed and map to \textsc{stop}, \textsc{slow} and \textsc{accel}.
The other three name a speed level that is maintained (slow, moderate or fast) and map to \textsc{keep}.

\paragraph{ORION.}
ORION~\cite{orion} outputs no speed token.
We read its yield decision by forced choice from the answer of its language head to the question ``Please describe your driving behavior and explain the reasons.''
We start the answer with ``You should'', and the stated decision is to yield if the first token of ``decelerate'' is more likely than the first tokens of ``keep'', ``maintain'' and ``accelerate'' together.

\paragraph{SimLingo.}
SimLingo~\cite{simlingo} states its decision as a commentary.
Every commentary template and each of its phrasings is cut at the first reason connector, and the remaining action phrase is assigned the decision of its canonical template.
The \prov{929} templates, each in \prov{21} phrasings, give \prov{160} distinct action phrases, each mapped to exactly one decision, and the mapping recovers the decision of all \prov{19{,}509} template instances.

\section{Implementation Details}
\label{supp:sec:impl}\suppanchor{\SuppParamsRef}{\thesection}

\subsection{Model Details}
\label{supp:sec:model}
\begin{table*}[t]
\centering
\caption{\textbf{Hyper-parameters of the model, the consistency losses and the training.} The online RL stage follows MindDrive~\cite{minddrive} and keeps its settings, except for the added progress reward.}
\label{supp:tab:hparams}
\small
\setlength{\tabcolsep}{4pt}
\begin{minipage}[t]{0.4825\linewidth}
\vspace{0pt}
\resizebox{\linewidth}{!}{%
\begin{tabular}{@{}ll@{}}
\toprule
Hyper-parameter & Value \\
\midrule
\multicolumn{2}{@{}l}{\emph{Model}} \\
\quad Vision encoder & EVA-02-L~\cite{eva02} \\
\quad Base LLM, hidden size & Qwen2-0.5B~\cite{qwen2}, 896 \\
\quad LoRA of each expert: rank, $\alpha$, dropout & 16, 16, 0.05 \\
\quad Speed waypoints $N_s$ & 6 (3\,s at 2\,Hz) \\
\midrule
\multicolumn{2}{@{}l}{\emph{Decision reading $\Phi$}} \\
\quad Stop threshold $\epsilon_v$ & 0.5\,m/s \\
\quad Speed-change threshold $\delta_v$ & 1\,m/s \\
\midrule
\multicolumn{2}{@{}l}{\emph{Commitment head}} \\
\quad Hidden size of the MLP for $\theta$ & 256 \\
\quad Value tokens: end speed, stopping distance & 16 (1\,m/s steps), 41 (2\,m steps) \\
\midrule
\multicolumn{2}{@{}l}{\emph{\codebook and fusion MLP}} \\
\quad Entries $\times$ dimension & $4\times64$ \\
\quad Fourier frequencies & 4 \\
\quad Hidden size of $\psi$ and of $G$ & 128 \\
\quad Output of the grounding head $G$ & $N_s$ speed values \\
\quad Hidden size of the fusion MLP $f$ & 512 \\
\quad Output layer of $f$ & zero-initialized \\
\midrule
\multicolumn{2}{@{}l}{\emph{Consistency losses}} \\
\quad Margin $m$ & 0.3\,m/s \\
\quad Weight $\lambda_p$ & 3 \\
\bottomrule
\end{tabular}}
\end{minipage}\hfill
\begin{minipage}[t]{0.4875\linewidth}
\vspace{0pt}
\resizebox{\linewidth}{!}{%
\begin{tabular}{@{}ll@{}}
\toprule
Hyper-parameter & Value \\
\midrule
\multicolumn{2}{@{}l}{\emph{IL, decision learning}} \\
\quad Optimizer & AdamW, weight decay $10^{-5}$ \\
\quad Learning rate of the trained parameters & $6.8\times10^{-5}$ \\
\quad Schedule & cosine over one epoch \\
\quad Warm-up & 500 iterations \\
\quad Batch size, training length & 6, 6{,}000 iterations \\
\quad Gradient clipping, precision & 35, FP16 \\
\midrule
\multicolumn{2}{@{}l}{\emph{IL, mapping learning}} \\
\quad Optimizer & AdamW, weight decay $10^{-5}$ \\
\quad Learning rate of the \codebook and $f$ & $3\times10^{-4}$ \\
\quad Learning rate of the speed branch $g$ & $3\times10^{-5}$ \\
\quad Schedule & cosine, 200 warm-up iterations \\
\quad Batch size, epochs & 256 frames, 40 \\
\quad Gradient clipping & 10 \\
\midrule
\multicolumn{2}{@{}l}{\emph{Online RL}} \\
\quad Discount factor, GAE parameter & 0.99, 1.0 \\
\quad Clip range & 0.2 \\
\quad Weights of the value, KL and entropy terms & 0.5, 0.5, 0 \\
\quad Optimizer & AdamW \\
\quad Learning rate of the LoRA parameters & $8\times10^{-6}$ \\
\quad Progress reward for a completed route & 1 \\
\quad Batch size & \prov{32} \\
\bottomrule
\end{tabular}}
\end{minipage}
\end{table*}

\method comprises a Decision Expert and an Action Expert, which share one VLM and differ only in their LoRA parameters.
We employ EVA-02-L~\cite{eva02} as the vision encoder and Qwen2-0.5B~\cite{qwen2} as the base LLM, which has a hidden size of 896.
Each expert has LoRA parameters of rank 16 on the query, key, value and output projections of the attention layers.
\cref{supp:tab:hparams} lists the hyper-parameters.
At inference, the decision is selected first, and the trajectory is generated under it (\cref{sec:method:problem}).
The Decision Expert outputs the commitment, the \codebook maps it to the code $z$, and the speed branch of the Action Expert generates the trajectory from its hidden state $h_a$ and the code (\cref{eq:code}).
The hidden state $h_a$ is computed with the planning prompt of MindDrive~\cite{minddrive} and does not depend on the commitment.

\paragraph{Decision Expert.}
The prompt of the Decision Expert carries the perception facts $F$ together with the current speed and the navigation command (\cref{supp:tab:templates}).
The Decision Expert states its decision in two forms, the commitment and the decision sentence, which are supervised by the same label (\cref{supp:sec:training}).
The commitment is read from the hidden state $h_c$ of the language model at the special token \texttt{<commit>}: a linear layer scores the four decisions from the normalized state, the decision is the admissible one with the highest score, and the target value is decoded for this decision:
\begin{equation}
\label{supp:eq:head}
\begin{aligned}
d&=\argmax_{k\in\Dadm}\,\big[W\,\mathrm{LN}(h_c)+b\big]_k,\\
\theta&=\mathrm{softplus}\Big(\mathrm{MLP}\big([\mathrm{LN}(h_c);\,\mathbf{1}_d;\,v_0/10]\big)\Big),
\end{aligned}
\end{equation}
where $\mathrm{LN}$ is layer normalization, $\mathbf{1}_d$ is the one-hot vector of $d$, and the MLP has one hidden layer of 256 units.
The target value is then clipped to the range of its decision with the margin $m$ of the consistency losses, for example to at least $v_0+\delta_v+m$ for \textsc{accel}.
At inference, $\Dadm$ is built from what the model observes.
It contains the decisions whose range of target values is not empty at the current speed, which excludes \textsc{slow} below 2.1\,m/s and \textsc{keep} below 0.1\,m/s.
These bounds are wider than those of the counterfactual executions (\cref{supp:sec:pipeline}), whose speed profiles are fixed.
\textsc{keep} and \textsc{accel} are also excluded when the perception facts report a red light that governs the ego lane.
The reach of \cref{supp:eq:reach} is not evaluated here, as the perception facts carry no distance to the stop line.
The admissible set of the labels additionally uses the logged future of the scene, which is not available at inference.
The decision sentence states this commitment in tokens, one for the decision and one for the target value (\cref{supp:tab:templates}), so the two forms state the same decision.

\paragraph{\codebook.}
The \codebook stores one learnable entry of dimension 64 for each of the four decisions, and the entries are initialized orthogonally.
The target value and the current speed are scaled to the order of one ($\theta$ by 20\,m for \textsc{stop} and by 10\,m/s otherwise, $v_0$ by 10\,m/s) and encoded with Fourier features~\cite{tancik2020fourier} with 4 frequencies:
\begin{equation}
\label{supp:eq:fourier}
\gamma(u)=\big[u,\ \sin(2^{i}\tfrac{\pi}{2}u),\ \cos(2^{i}\tfrac{\pi}{2}u)\big]_{i=0}^{3}.
\end{equation}
The MLP $\psi$ takes the one-hot vector of $d$ together with $\gamma(\theta)$ and $\gamma(v_0)$, and its output is added to the entry $\mathbf{E}[d]$, which forms the commitment code $z$ (\cref{eq:code}).
The grounding head $G$ takes $z$ and $v_0$, the latter as $\gamma(v_0)$, and decodes the nominal speed profile $\hat w$ of $N_s=6$ values.
Both $\psi$ and $G$ have two hidden layers of 128 units.

\paragraph{Fusion and speed branch.}
The fusion MLP $f$ has two hidden layers of 512 units.
It reads the hidden state $h_a$ after a layer normalization, the code $z$, the profile $\hat w$ and the speed $v_0$, where $\hat w$ enters without a gradient.
Its output layer is initialized to zero, so training starts from the output $g(h_a)$ of the pretrained speed branch, and the commitment enters as a residual.
The speed branch $g$ keeps the architecture of MindDrive~\cite{minddrive} (a prior network, a GRU and a decoder of seven candidate trajectories) and is initialized from the released weights.
For each decision, the trajectory is taken from one fixed candidate of the decoder, that of the speed token \texttt{stop}, \texttt{slow\_down}, \texttt{maintain\_moderate\_speed} or \texttt{speed\_up} of MindDrive, respectively.
The latent variable of the prior network is set to its mean.
The commitment enters through the fused state, which all candidates of the decoder share.
The choice of the candidate alone does not make the trajectory follow the decision: the candidates of MindDrive follow their own speed token on 27\% to 30\% of the frames, and with the speed branch frozen at the weights of MindDrive, the trajectory of \method still follows every forced decision on 98.8\% of the frames (\cref{supp:sec:intervention,supp:sec:ablation}).
Together with the \codebook, the branch has 16.0M parameters.
The path branch of the Action Expert stays frozen, and the navigation command selects its path as in MindDrive.
On all 12{,}806 validation frames, the path token that the released checkpoint predicts equals the navigation command.
The open-loop error also comes almost entirely from the speed.
On the validation subset (\cref{supp:sec:protocol}), the error of the released checkpoint falls from 1.19\,m to 0.16\,m when the speed profile of the expert is placed on the planned path, and it stays at 1.14\,m when the planned speed profile is placed on the path of the expert.

\paragraph{Reading a trajectory.}
The consistency losses of \cref{eq:cons} read three quantities from a trajectory $a$: its speed profile $w(a)=(v_1,\dots,v_{N_s})$ of step speeds (\cref{supp:sec:pipeline}), its target value $\Theta(a)$ of \cref{supp:eq:theta}, and the signed distance $\rho_k$ of its speed profile to the boundary of decision $k$ under $\Phi$.
For a given current speed $v_0$, \cref{supp:eq:phi} bounds the end speed of each decision by one or two conditions, with the signed distances
\begin{equation}
\label{supp:eq:rho}
\begin{aligned}
\textsc{stop}:\quad&\epsilon_v-v_{N_s},\\
\textsc{slow}:\quad&v_0-v_{N_s}-\delta_v,\quad v_{N_s}-\epsilon_v,\\
\textsc{keep}:\quad&\delta_v-|v_{N_s}-v_0|,\quad v_{N_s}-\epsilon_v,\\
\textsc{accel}:\quad&v_{N_s}-v_0-\delta_v,\quad v_{N_s}-\epsilon_v,
\end{aligned}
\end{equation}
and $\rho_k$ is the smallest distance of decision $k$.
A distance is positive when its condition holds, and it is measured in m/s.
With the margin $m=0.3$\,m/s, the hinge $[m-\rho_k]_+$ requires the profile to lie at least 0.3\,m/s inside decision $k$.
We implement it as the sum of the hinges of the individual conditions, which equals $[m-\rho_k]_+$ whenever at most one of them is active.

\subsection{Training Details}
\label{supp:sec:training}
\begin{algorithm}[t]
\caption{\textbf{One training step of the consistent language-action mapping}, written for one frame of the batch.}
\label{supp:alg:training}
\begin{algorithmic}[1]
\Require frame $(V,L)$ with current speed $v_0$, admissible set $\Dadm$ and targets $\{\bar a_k\}_{k\in\Dadm}$
\Trainable \codebook $(\mathbf{E},\psi,G)$, fusion MLP $f$, speed branch $g$; all other modules frozen
\State $h_a\gets$ hidden state of the Action Expert for $(V,L)$, cached
\ForAll{$k\in\Dadm$}
  \Phase{$\bar a_k$: expert trajectory if $k=\bar d$, else counterfactual execution (one exception, \cref{supp:sec:training})}
  \State $\theta_k\gets\Theta(\bar a_k)$; \ $\tilde\theta_k\gets\theta_k$ with noise, clipped to the range of $k$
  \State $z_k\gets\mathbf{E}[k]+\psi\big(k,\gamma(\tilde\theta_k),\gamma(v_0)\big)$
  \State $\hat w_k\gets G(z_k,v_0)$
  \State $a_k\gets g\big(h_a+f([h_a;z_k;\hat w_k;v_0])\big)$
  \State compute $\mathcal{L}_{\mathrm{exec}}$ from $a_k$, $\bar a_k$ and $\theta_k$, and $\mathcal{L}_{\mathrm{ground}}$ and $\mathcal{L}_{\mathrm{sem}}$ from the profile that $G$ decodes for $\theta_k$
\EndFor
\State $\mathcal{L}_{\mathrm{cons}}\gets\frac{1}{|\Dadm|}\sum_{k\in\Dadm}\big(\mathcal{L}_{\mathrm{ground}}+\mathcal{L}_{\mathrm{sem}}+\mathcal{L}_{\mathrm{exec}}\big)$
\State update $\mathbf{E}$, $\psi$, $G$, $f$ and $g$ by back-propagation on $\mathcal{L}_{\mathrm{cons}}$
\end{algorithmic}
\end{algorithm}

\method is trained in two stages, IL and online RL, both following MindDrive~\cite{minddrive}.
The IL stage consists of decision learning and mapping learning.

\paragraph{Decision learning.}
The Action Expert is initialized from the released checkpoint of MindDrive, in which it has been trained by IL alone.
The vocabulary is extended by 62 new tokens: \texttt{<commit>}, the four decision tokens and the 57 value tokens of \cref{supp:tab:templates}.
In decision learning, the LoRA parameters of the Decision Expert, the commitment head and the embeddings of the new tokens are trained, and all other parameters are frozen, so $h_a$ and the path are unchanged.
The Decision Expert is trained with three losses on the labeled frames.
The language modeling loss supervises the decision sentence.
The commitment loss supervises the scores $q=\mathrm{softmax}(W\,\mathrm{LN}(h_c)+b)$ of the commitment head, which implement the selection policy $\pi_d$, with the label $\bar d$ and with the admissible set:
\begin{equation}
\label{supp:eq:commit}
\mathcal{L}_{\mathrm{commit}}=-\log q_{\bar d}-\log\sum_{k\in\Dadm}q_k .
\end{equation}
The target value that the head decodes for $\bar d$ is supervised by an L1 loss with respect to $\bar\theta$.
The label $(\bar d,\bar\theta)$ is that of \cref{supp:sec:pipeline}, and the prompt carries the perception facts of the frame.
If the decision of the expert is not admissible, because the expert keeps its speed or accelerates where a traffic rule forbids it, the label is replaced by \textsc{stop} with the target value of its counterfactual execution.
This affects 2.4\% of the labeled training frames.
If no feasible \textsc{stop} exists either, only the second term of \cref{supp:eq:commit} is applied.

\paragraph{Mapping learning.}
The \codebook, the fusion MLP and the speed branch are then trained with the consistency loss $\mathcal{L}_{\mathrm{cons}}$ of \cref{eq:consall}, while all other modules are frozen.
The VLM is among the frozen modules, so the hidden states $h_a$ are computed once and cached.
The speed branch of the reported results is trained on every third training clip, 317 clips with 78{,}078 frames, of which 65{,}398 carry a label, while the Decision Expert is trained on frames of all training clips.
\cref{supp:alg:training} gives one training step.
For every admissible decision $k$ of a frame, the commitment $(k,\theta_k)$ is read from the target $\bar a_k$, which is the expert trajectory for the decision of the expert and a counterfactual execution otherwise, with one exception that is given below.
One frame contributes up to four targets, 2.7 on average, which differ only in the decision.
The \codebook is fitted alone beforehand, with $\mathcal{L}_{\mathrm{ground}}$ and $\mathcal{L}_{\mathrm{sem}}$ on the 618{,}896 valid executions of the labeled training frames, expert and counterfactual.
The target value that the \codebook receives is perturbed by Gaussian noise with a standard deviation of 1.5\,m for \textsc{stop} and of 2.0, 0.5 and 1.0\,m/s for \textsc{slow}, \textsc{keep} and \textsc{accel}, and clipped to the range of its decision.
This imitates the error of the Decision Expert, and $\mathcal{L}_{\mathrm{ground}}$ uses the value without noise.
The exception concerns the trajectory that the pretrained speed branch generates without a commitment.
If $\Phi$ reads this trajectory as a decision other than that of the expert, it replaces the counterfactual execution of that decision as the target, and the pair is left out of $\mathcal{L}_{\mathrm{ground}}$ and $\mathcal{L}_{\mathrm{sem}}$.
The trajectory term of $\mathcal{L}_{\mathrm{exec}}$ is the L1 error of the positions plus half the L1 error of the displacements between consecutive waypoints.

\paragraph{Online RL.}
The second stage follows the online RL of MindDrive~\cite{minddrive}: rollouts are collected in CARLA~\cite{carla}, and the Decision Expert is updated by proximal policy optimization (PPO) on the collected steps.
The RL action is the decision of the commitment.
It is sampled from the scores of the commitment head over $\Dadm$, its target value is decoded as in \cref{supp:eq:head}, and the frozen Action Expert executes it, so the stated decision is also the one that the Action Expert executes during exploration.
The reward is that of MindDrive, $-1$ for a penalty event, $+1$ for a completed route and 0 otherwise.
We add a progress reward that is paid in proportion to the route completion and amounts to 1 for a completed route.
A penalty event ends the episode.
The seven penalty events are a collision with a pedestrian, with a vehicle or with a static object; a traffic-light or a stop-sign infraction; a deviation from the route; and leaving the lanes of the route.
As in MindDrive, PPO updates only the LoRA parameters of the Decision Expert and the output layer of the value network, which we initialize to zero.

\subsection{Formal Statement on Expert-Only Supervision}
\label{supp:sec:shortcut}\suppanchor{\SuppFormalRef}{\thesubsection}

\cref{sec:method:mapping} states that a conditioned planner can ignore its condition.
Let $x=(V,L)$ denote the input of a frame.
Expert data contains one trajectory $a^{\mathrm{e}}(x)$ per frame, so the commitment that is read from it, $c^{\mathrm{e}}(x)=\big(\Phi(a^{\mathrm{e}}(x)),\Theta(a^{\mathrm{e}}(x))\big)$, is a function of $x$.
We take the generation policy $\pi_g$ as a deterministic planner $P(x,c)$ that maps an input and a commitment to a trajectory.
Imitation on expert data minimizes
\begin{equation}
\label{supp:eq:imitation}
J(P)=\mathbb{E}_{x}\Big[\ell\big(P(x,c^{\mathrm{e}}(x)),\,a^{\mathrm{e}}(x)\big)\Big]
\end{equation}
for a loss $\ell$.

\begin{proposition}
\label{supp:prop:shortcut}
(i) For every planner $P$, there is a planner $P'$ that ignores the commitment and has $J(P')=J(P)$.
The minimum of $J$ over the planners that ignore the commitment therefore equals its minimum over all planners.
(ii) Two planners that agree on all pairs $(x,c^{\mathrm{e}}(x))$ have the same value of $J$, whatever they output for a commitment $c\neq c^{\mathrm{e}}(x)$.
\end{proposition}

\noindent\emph{Proof.}
For (i), let $P'(x,c)=P(x,c^{\mathrm{e}}(x))$.
The planner $P'$ does not depend on $c$ and agrees with $P$ on all pairs $(x,c^{\mathrm{e}}(x))$, so $J(P')=J(P)$.
For (ii), \cref{supp:eq:imitation} evaluates a planner only on the pairs $(x,c^{\mathrm{e}}(x))$. \qed

\vspace{.5em}
Under expert-only supervision, the objective therefore neither rewards the use of the decision nor constrains the response of the planner to a decision other than that of the expert.
This agrees with the follow rate of MindDrive (\cref{supp:sec:intervention}) and with the ablation of \cref{supp:sec:ablation}, in which a branch that is supervised only on the decision of the expert follows a forced decision far less reliably.
The consistency loss of \method addresses both points: for an admissible decision $k$ other than that of the expert, the pair $(x,(k,\theta_k))$ enters the objective with the target $\bar a_k$.
This target differs from $a^{\mathrm{e}}(x)$, since $\Phi(\bar a_k)=k$.
A planner that ignores the commitment cannot fit both targets, and the response of the planner to the other admissible decisions is supervised.

\subsection{Evaluation Protocol}
\label{supp:sec:protocol}
\begin{table}[t]
\centering
\caption{\textbf{Routes of Dev10}, the development subset of DriveTransformer~\cite{drivetransformer}.}
\label{supp:tab:dev10}
\small
\setlength{\tabcolsep}{5pt}
\begin{tabular}{@{}rll@{}}
\toprule
Route & Scenario & Town \\
\midrule
2091 & NonSignalizedJunctionLeftTurn & Town12 \\
3255 & ParkingCrossingPedestrian & Town13 \\
3514 & ParkingExit & Town13 \\
17569 & SequentialLaneChange & Town12 \\
25378 & YieldToEmergencyVehicle & Town03 \\
25381 & HazardAtSideLane & Town05 \\
25424 & ConstructionObstacleTwoWays & Town11 \\
26405 & StaticCutIn & Town15 \\
27494 & BlockedIntersection & Town04 \\
28198 & SignalizedJunctionLeftTurnEnterFlow & Town15 \\
\bottomrule
\end{tabular}
\end{table}

\paragraph{Closed loop.}
The closed-loop results that we measure ourselves ($^\ddagger$ in \cref{tab:main}) are obtained with the official Bench2Drive evaluator~\cite{bench2drive} on its 220 routes in CARLA \prov{0.9.15}~\cite{carla}, in synchronous mode at \prov{20}\,Hz.
The evaluation runs on a single RTX 4090 with 24\,GB of memory, which holds the simulator and the model together.
The baselines are run from their released checkpoints with their official agents.
One route, a highway exit in Town12, crashes the simulator for MindDrive and for ORION.
It counts as zero in their scores, and their consistency analysis covers the other 219 routes.
Each of these results is the median over multiple random seeds.
For the released MindDrive checkpoint, \prov{149} of the \prov{219} scored routes receive the same score in two runs and \prov{27} change their success flag, which puts the standard deviation of a single run at about \prov{1.0} points of DS and \prov{1.7} points of SR.

\paragraph{Dev10.}
Dev10 is the development subset of DriveTransformer~\cite{drivetransformer}, which contains ten of the 220 routes and is distributed with the Bench2Drive evaluator (\cref{supp:tab:dev10}).
It is used for the say-do gap of \method (\cref{tab:gap}), for the rule-based decisions and for the latency breakdown (\cref{tab:rule}).
\cref{supp:sec:gapfull} also evaluates the baselines on these ten routes.

\paragraph{Consistency logging.}
For the consistency analysis, every tenth simulator tick is logged, which gives a rate of 2\,Hz.
A logged frame contains the detections of the model, its stated decision, its planned trajectory, the trajectory executed over the following horizon and the privileged state of all agents within \prov{30}\,m.
The loggers do not affect the run: replaying logged ticks with and without them gives identical plans and controls.

\paragraph{Consistency metrics.}
The say-do gap of \cref{eq:saydo} compares the stated decision $d_t$ with the decision $\Phi(a_t)$ of the planned trajectory.
For \method, the stated decision is the decision of the commitment, which the decision sentence states (\cref{supp:sec:model}).
The say-exec gap replaces $a_t$ by the trajectory that the vehicle executes over the same horizon, and the do-exec gap compares the planned with the executed trajectory.
Both leave out the last frames of a route, for which the executed horizon is incomplete.
Each gap is reported for the speed decision and for the yield decision of \cref{supp:sec:mapping}.
In closed loop, the trajectories of all models are read over a horizon of \prov{2.5}\,s, the longest that every evaluated model plans (\cref{supp:sec:gapfull}), on all logged frames.
The commitment of \method is defined over 3\,s, and read over 3\,s, its say-do gap on Dev10 is 0.0\% (0 of 1{,}769 frames) instead of 1.8\%.
In open loop, all trajectories are read over 3\,s.
The decisions are imbalanced, so a model that always states \textsc{keep} and mostly drives at a constant speed has a small gap without any dependence between its two outputs.
We therefore also report a normalized gap, which divides the gap by the one expected from two independent outputs with the same marginals:
\begin{equation}
\label{supp:eq:norm}
\widetilde{\mathrm{SD}}=\frac{\mathrm{SD}}{1-\sum_{k\in\Dset}p_{\mathrm{say}}(k)\,p_{\mathrm{do}}(k)}=1-\kappa,
\end{equation}
where $p_{\mathrm{say}}(k)$ and $p_{\mathrm{do}}(k)$ are the shares of frames on which the stated decision and the decision of the trajectory, respectively, equal $k$, and $\kappa$ is Cohen's coefficient of agreement~\cite{cohen1960kappa}.
A value of 0 means perfect agreement, and a value of 1 means that the two outputs agree no more often than independent ones.
A value above 1 means that they agree less often.

\paragraph{Decision interventions.}
The gaps are correlational, so we also test whether the decision determines the trajectory: we replay logged frames, force the decision to a value $k$, decode the trajectory again and report the follow rate
\begin{equation}
\label{supp:eq:follow}
\mathrm{Follow}(k)=\frac{1}{N_f}\sum_{t=1}^{N_f}\mathbb{I}\Big[\Phi\big(a_t^{(k)}\big)=k\Big],
\end{equation}
where $a_t^{(k)}$ is the trajectory that the model generates at frame $t$ under the forced decision $k$.
The forced decision is a speed token for MindDrive, a commentary for SimLingo and a commitment for \method.
For MindDrive, $a_t^{(k)}$ is the candidate trajectory of the forced speed token, the token of decision $k$ that \cref{supp:sec:model} names.
For \method, the target value of the forced commitment is read from the execution of $k$ in the same frame.
For both models, the sum runs over the frames on which this execution exists, including those on which $k$ is the decision of the expert.

\paragraph{Open loop.}
Open-loop results use the 50 validation clips with 12{,}806 frames, of which 10{,}806 carry a label.
The ablations of \cref{supp:sec:ablation} use the validation subset, the first 80 frames of every validation clip.
It has 3{,}987 frames, since one clip has only 67, and 3{,}434 of them carry a label.
The decision interventions on MindDrive and \method use this subset and all validation frames, and those on SimLingo use its closed-loop logs (\cref{supp:sec:intervention}).
The average L2 of \cref{tab:main} is taken over 2\,s at 2\,Hz as in UniAD~\cite{uniad}, and the L2 of \cref{tab:ablation} is the error at 3\,s.
The trajectory errors and collision rates of this supplementary material are computed with the evaluation code of MindDrive~\cite{minddrive} in the protocol of ST-P3~\cite{stp3}.

\paragraph{Confidence intervals.}
Where intervals are given, they are \prov{95\%} percentile intervals of a bootstrap over routes with \prov{1{,}000} resamples.
One sampling matrix is shared by all models, so that differences are paired.

\section{More Results}
\label{supp:sec:results}
\setcounter{dbltopnumber}{3}

\subsection{Runtime Analysis}
\label{supp:sec:runtime}
\begin{table}[t]
\centering
\caption{\textbf{Runtime of \method and of existing VLA models.} Upper block: mean model time per step on an NVIDIA A100, as reported by MindDrive~\cite{minddrive}. Middle block: model time per step on a single RTX 4090, without simulator and control (totals of \cref{tab:rule}). Lower block: median wall-clock time of one closed-loop step on a single RTX 4090, with the simulator tick included (\cref{supp:sec:runtime}). Reproduced numbers of the baselines in italics.}
\label{supp:tab:runtime}
\small
\setlength{\tabcolsep}{4pt}
\resizebox{\linewidth}{!}{%
\begin{tabular}{@{}llcc@{}}
\toprule
Method & LLM & GPU & Time (ms)\,$\downarrow$ \\
\midrule
\multicolumn{4}{@{}l}{\emph{Model time, as reported by MindDrive~\cite{minddrive}}} \\
ORION~\cite{orion} & Vicuna-7B & A100 & $\sim$1{,}106 \\
ReCogDrive~\cite{recogdrive} & Qwen2.5-7B & A100 & $\sim$750 \\
MindDrive~\cite{minddrive} & Qwen2-0.5B & A100 & $\sim$540 \\
\midrule
\multicolumn{4}{@{}l}{\emph{Model time per step, measured}} \\
ORION~\cite{orion} & Vicuna-7B & RTX 4090 & \repro{251.1} \\
MindDrive~\cite{minddrive} & Qwen2-0.5B & RTX 4090 & \repro{246.2} \\
\textbf{\method} & Qwen2-0.5B & RTX 4090 & 259.8 \\
\quad + rule-triggered & Qwen2-0.5B & RTX 4090 & 216.7 \\
\midrule
\multicolumn{4}{@{}l}{\emph{Closed-loop step, simulator included}} \\
MindDrive, 220 routes & Qwen2-0.5B & RTX 4090 & \repro{582} \\
MindDrive, two routes & Qwen2-0.5B & RTX 4090 & \repro{588} \\
\textbf{\method}, two routes & Qwen2-0.5B & RTX 4090 & 547 \\
\quad + rule-triggered & Qwen2-0.5B & RTX 4090 & 474 \\
\bottomrule
\end{tabular}}
\end{table}

We analyze the runtime of \method on a single RTX 4090.
\cref{tab:rule} gives the model time per step, which is measured inside the agent and excludes the simulator and the control.
Perception, which comprises the vision encoder and the perception heads, takes 189.8 of the 259.8\,ms of \method.
The Decision Expert takes 35.5\,ms and the planner 33.1\,ms (MindDrive: 34.5\,ms and 26.3\,ms).
When a rule issues the decision, the decision takes 5.4\,ms instead of 35.5\,ms, and the total falls to 216.7\,ms.
\cref{supp:tab:runtime} places these totals next to the runtimes that MindDrive~\cite{minddrive} reports on an NVIDIA A100 and next to the wall-clock time of one closed-loop step.
A closed-loop step is the interval between two consecutive calls of the agent by the evaluator.
It contains one simulator tick with six cameras and one forward pass of the model.
For MindDrive, a step takes a median of \prov{582}\,ms and a mean of \prov{647}\,ms over its 220-route evaluation.
On routes 17569 and 28198 of Dev10, on which the model times of MindDrive and \method in \cref{tab:rule} are measured, the median step time is 547\,ms for \method and 588\,ms for MindDrive.
On the steps of Dev10 on which a rule issues the decision, it is 474\,ms.

\subsection{Consistency of All Evaluated Models}
\label{supp:sec:gapfull}\suppanchor{\SuppAnalysisRef}{\thesubsection}
\begin{table*}[t]
\centering
\caption{\textbf{Models of the consistency analysis.} The baselines are evaluated from their released checkpoints. Plan: waypoints $\times$ interval of the planned trajectory. Stated decision: the output that is mapped to $\Dset$ (\cref{supp:sec:mapping}). Frames, Routes: logged closed-loop frames at 2\,Hz and the routes on which they are logged.}
\label{supp:tab:models}
\small
\setlength{\tabcolsep}{6pt}
\begin{tabular}{@{}lllllrr@{}}
\toprule
Method & LLM & Plan & Checkpoint & Stated decision & Frames & Routes \\
\midrule
ORION~\cite{orion} & Vicuna-7B, LoRA & $6\times0.5$\,s & released, FP16 & yield decision, forced choice & 26{,}638 & 219 \\
MindDrive~\cite{minddrive} & Qwen2-0.5B, two LoRAs & $6\times0.5$\,s & released, online RL & speed token & 24{,}317 & 219 \\
SimLingo~\cite{simlingo} & InternVL2-1B & $10\times0.25$\,s & released, epoch 13 & commentary & 19{,}149 & 220 \\
\textbf{\method} & Qwen2-0.5B, two LoRAs & $6\times0.5$\,s & ours, after IL & commitment & 1{,}769 & 10 \\
\bottomrule
\end{tabular}
\end{table*}

\begin{table*}[t]
\centering
\caption{\textbf{Say-do consistency of all evaluated models in closed loop on Bench2Drive.} Raw: share of frames (\%) on which the stated decision differs from the decision of the planned trajectory (say-do) or of the executed trajectory (say-exec), as in \cref{tab:gap}. $1{-}\kappa$: the normalized gap of \cref{supp:eq:norm}. Lon.: speed decision; Yield: yield decision. ORION states only the yield decision. Frames: logged frames. Upper block: the baselines on all their logged routes (\cref{supp:tab:models}), as in \cref{tab:gap}. Lower block: all models on the ten routes of Dev10 (\cref{supp:tab:dev10}).}
\label{supp:tab:gapfull}
\small
\setlength{\tabcolsep}{5pt}
\begin{tabular}{@{}lrcccccccc@{}}
\toprule
 & & \multicolumn{2}{c}{Say-do, Lon.} & \multicolumn{2}{c}{Say-do, Yield} & \multicolumn{2}{c}{Say-exec, Lon.} & \multicolumn{2}{c}{Say-exec, Yield} \\
\cmidrule(lr){3-4}\cmidrule(lr){5-6}\cmidrule(lr){7-8}\cmidrule(l){9-10}
Method & Frames & Raw\,$\downarrow$ & $1{-}\kappa\,\downarrow$ & Raw\,$\downarrow$ & $1{-}\kappa\,\downarrow$ & Raw\,$\downarrow$ & $1{-}\kappa\,\downarrow$ & Raw\,$\downarrow$ & $1{-}\kappa\,\downarrow$ \\
\midrule
\multicolumn{10}{@{}l}{\emph{All logged routes}} \\
ORION~\cite{orion} & 26{,}638 & -- & -- & 23.8 & 0.97 & -- & -- & 65.7 & 1.00 \\
MindDrive~\cite{minddrive} & 24{,}317 & 83.5 & 0.95 & 19.8 & 0.84 & 69.6 & 0.94 & 59.9 & 0.98 \\
SimLingo~\cite{simlingo} & 19{,}149 & 49.3 & 0.69 & 43.4 & 0.78 & 21.0 & 0.41 & 16.1 & 0.42 \\
\midrule
\multicolumn{10}{@{}l}{\emph{Ten routes of Dev10}} \\
ORION~\cite{orion} & 1{,}339 & -- & -- & 29.4 & 0.98 & -- & -- & 73.3 & 1.02 \\
MindDrive~\cite{minddrive} & 1{,}582 & 80.1 & 0.96 & 18.8 & 0.80 & 80.2 & 0.97 & 69.7 & 0.99 \\
SimLingo~\cite{simlingo} & 904 & 59.6 & 0.77 & 51.9 & 0.88 & 19.4 & 0.36 & 14.3 & 0.36 \\
\textbf{\method} & 1{,}769 & 1.8 & 0.03 & 0.8 & 0.02 & 52.4 & 0.82 & 39.8 & 0.74 \\
\bottomrule
\end{tabular}
\end{table*}

\begin{table*}[t]
\centering
\caption{\textbf{More ablations of the consistent language-action mapping in open loop}, on the 3{,}434 labeled frames of the validation subset (\cref{supp:sec:protocol}), with measures that \cref{tab:ablation} does not report. Each variant retrains the \codebook, the fusion MLP and, unless frozen, the speed branch. Col.: collision rate at 3\,s in the protocol of ST-P3~\cite{stp3}, with the label of the frame as the commitment (label) and with the commitment that a Decision Expert outputs (predicted), here an earlier checkpoint than the reported one. Sep.: mean distance between the end points of the trajectories of two decisions in the same frame. All: share of frames on which the trajectory follows every forced decision (\cref{supp:eq:follow}). Runs: number of random seeds; the mean over the runs is given, and the reported model is one of the three runs of the first row.}
\label{supp:tab:ablation}
\small
\setlength{\tabcolsep}{8pt}
\begin{tabular}{@{}lccccc@{}}
\toprule
 & & \multicolumn{2}{c}{Col. (\%)\,$\downarrow$} & & \\
\cmidrule(lr){3-4}
Variant & Runs & Label & Predicted & Sep. (m) & All (\%)\,$\uparrow$ \\
\midrule
\textbf{\method} & 3 & 0.50 & 1.31 & 8.39 & 99.0 \\
\midrule
w/o counterfactual data & 1 & 1.08 & 1.16 & 2.65 & 32.4 \\
w/o $\mathcal{L}_{\mathrm{ground}}$ and $\mathcal{L}_{\mathrm{sem}}$ & 3 & 0.52 & 1.28 & 8.39 & 99.1 \\
speed branch $g$ frozen & 3 & 0.60 & 1.33 & 8.43 & 98.8 \\
20 instead of 40 epochs & 1 & 0.49 & 1.21 & 8.41 & 98.8 \\
\bottomrule
\end{tabular}
\end{table*}

\begin{table*}[t]
\begin{minipage}[t]{\columnwidth}
\centering
\caption{\textbf{Decision interventions on the labeled validation frames.} Subset: the 3{,}434 labeled frames of the validation subset, which give the numbers behind \cref{fig:teaser}b. All: the 10{,}806 labeled frames of the validation set. Follow: share of frames (\%) on which the trajectory under a forced decision is read back as this decision (\cref{supp:eq:follow}). Sep.: mean distance between the end points of the trajectories of two forced decisions in the same frame. A decision is forced on the frames that have an execution of it: 3{,}414, 2{,}509, 2{,}666 and 3{,}429 frames of the subset and 10{,}741, 5{,}340, 5{,}934 and 10{,}795 of all frames, in the order of the columns.}
\label{supp:tab:follow}
\small
\setlength{\tabcolsep}{3.5pt}
\begin{tabular}{@{}llccccc@{}}
\toprule
 & & \multicolumn{4}{c}{Follow (\%)\,$\uparrow$} & Sep. \\
\cmidrule(lr){3-6}
Frames & Method & \textsc{stop} & \textsc{slow} & \textsc{keep} & \textsc{accel} & (m) \\
\midrule
Subset & MindDrive~\cite{minddrive} & 27.4 & 29.7 & 30.3 & 27.7 & 0.51 \\
 & \textbf{\method} & 99.9 & 99.9 & 99.4 & 99.4 & 8.38 \\
\midrule
All & MindDrive~\cite{minddrive} & 35.5 & 24.6 & 34.4 & 33.8 & 0.63 \\
 & \textbf{\method} & 100.0 & 99.9 & 99.2 & 99.1 & 8.20 \\
\bottomrule
\end{tabular}

\end{minipage}\hfill
\begin{minipage}[t]{\columnwidth}
\centering
\caption{\textbf{Interventions on SimLingo in closed loop}, on 9{,}620 of its logged frames from all 220 routes. Forced commentary: the commentary is set to a decision $k$ and the trajectory is decoded again; follow as in \cref{supp:eq:follow}. $\Delta v$: difference in end speed between two forced decisions. Forced instruction: the instruction-following task that Action Dreaming trains, on 4{,}865 of these frames. Resampling: a second forward pass with the dropout of the model active, with the share of changed decisions of the commentary and of the plan. Removing the commentary: the trajectory is decoded without the commentary, and the row gives how often the decision of the plan changes and the mean change of its end speed. Brackets: \prov{95\%} route-bootstrap intervals.}
\label{supp:tab:probes}
\small
\setlength{\tabcolsep}{4pt}
\begin{tabular}{@{}lc@{}}
\toprule
Intervention & Value \\
\midrule
\multicolumn{2}{@{}l}{\emph{Forced commentary}} \\
\quad follow (\%), \textsc{stop} / \textsc{slow} / \textsc{keep} / \textsc{accel} & \prov{29} / \prov{13} / \prov{20} / \prov{67} \\
\quad follow (\%), average & \prov{32} [\prov{31}, \prov{33}] \\
\quad $\Delta v$, \textsc{accel} vs.\ \textsc{slow} & \prov{+3.1}\,m/s [\prov{2.6}, \prov{3.7}] \\
\quad $\Delta v$, \textsc{keep} vs.\ \textsc{stop} & \prov{+0.5}\,m/s [\prov{0.4}, \prov{0.6}] \\
\midrule
\multicolumn{2}{@{}l}{\emph{Forced instruction}} \\
\quad success (\%), stop / slower / faster & \prov{99.6} / \prov{94.1} / \prov{99.9} \\
\midrule
\multicolumn{2}{@{}l}{\emph{Resampling with dropout}} \\
\quad changed decision (\%), commentary / plan & \prov{4.6} / \prov{7.3} \\
\midrule
\multicolumn{2}{@{}l}{\emph{Removing the commentary}} \\
\quad changed decision (\%) / mean $|\Delta v|$ & \prov{13.9} / \prov{0.72}\,m/s \\
\bottomrule
\end{tabular}

\end{minipage}
\end{table*}

\begin{table*}[t]
\centering
\caption{\textbf{Gaps and infractions of existing models by scene type}, in closed loop, for the yield decision. S-D: say-do gap; S-E: say-exec gap; D-E: do-exec gap, all in \% of the frames. Infr.: infractions per kilometer on the frames of the scene type. A frame can belong to several complex scene types. For SimLingo, 19{,}147 of the 19{,}149 logged frames have a stated decision.}
\label{supp:tab:strata}
\small
\setlength{\tabcolsep}{4.6pt}
\begin{tabular}{@{}l rcccc rcccc rcccc@{}}
\toprule
 & \multicolumn{5}{c}{MindDrive~\cite{minddrive}} & \multicolumn{5}{c}{ORION~\cite{orion}} & \multicolumn{5}{c}{SimLingo~\cite{simlingo}} \\
\cmidrule(lr){2-6}\cmidrule(lr){7-11}\cmidrule(l){12-16}
Scene type & Frames & S-D & S-E & D-E & Infr. & Frames & S-D & S-E & D-E & Infr. & Frames & S-D & S-E & D-E & Infr. \\
\midrule
Simple & \prov{1{,}609} & \prov{6.0} & \prov{13.2} & \prov{10.1} & \prov{4.8} & \prov{1{,}479} & \prov{5.3} & \prov{9.1} & \prov{5.4} & \prov{3.6} & \prov{471} & \prov{18.7} & \prov{28.1} & \prov{23.7} & \prov{1.8} \\
C1 junction or turn & \prov{10{,}869} & \prov{26.2} & \prov{66.7} & \prov{47.1} & \prov{14.0} & \prov{11{,}838} & \prov{29.9} & \prov{73.5} & \prov{48.1} & \prov{12.9} & \prov{10{,}157} & \prov{43.9} & \prov{16.3} & \prov{49.9} & \prov{6.1} \\
C2 vulnerable road user & \prov{3{,}344} & \prov{25.5} & \prov{52.0} & \prov{36.1} & \prov{5.0} & \prov{3{,}810} & \prov{31.6} & \prov{62.8} & \prov{38.6} & \prov{7.8} & \prov{2{,}952} & \prov{42.7} & \prov{17.5} & \prov{43.2} & \prov{2.2} \\
C3 conflict or cut-in & \prov{9{,}830} & \prov{15.6} & \prov{68.6} & \prov{56.8} & \prov{14.2} & \prov{11{,}984} & \prov{16.1} & \prov{73.5} & \prov{59.7} & \prov{12.8} & \prov{7{,}547} & \prov{44.0} & \prov{20.1} & \prov{51.8} & \prov{7.0} \\
C4 stopped or low speed & \prov{15{,}612} & \prov{26.7} & \prov{82.2} & \prov{60.9} & \prov{27.4} & \prov{18{,}137} & \prov{31.7} & \prov{87.8} & \prov{60.4} & \prov{28.1} & \prov{14{,}030} & \prov{52.0} & \prov{12.4} & \prov{56.1} & \prov{8.3} \\
C5 static obstacle & \prov{5{,}890} & \prov{17.3} & \prov{59.7} & \prov{45.5} & \prov{5.3} & \prov{8{,}300} & \prov{21.3} & \prov{74.1} & \prov{53.9} & \prov{3.9} & \prov{5{,}134} & \prov{42.5} & \prov{13.7} & \prov{46.1} & \prov{3.6} \\
C6 night or rain & \prov{13{,}702} & \prov{19.0} & \prov{61.5} & \prov{47.7} & \prov{9.1} & \prov{14{,}537} & \prov{22.2} & \prov{66.3} & \prov{47.8} & \prov{7.7} & \prov{10{,}211} & \prov{46.2} & \prov{17.4} & \prov{50.7} & \prov{3.8} \\
C7 dense interaction & \prov{8{,}050} & \prov{16.2} & \prov{40.9} & \prov{29.5} & \prov{8.6} & \prov{9{,}305} & \prov{18.0} & \prov{48.7} & \prov{34.9} & \prov{9.1} & \prov{7{,}155} & \prov{37.2} & \prov{21.1} & \prov{45.1} & \prov{6.4} \\
\midrule
Complex (any) & \prov{22{,}694} & \prov{20.8} & \prov{63.3} & \prov{47.6} & \prov{9.8} & \prov{25{,}152} & \prov{24.9} & \prov{69.1} & \prov{47.9} & \prov{8.7} & \prov{18{,}673} & \prov{44.1} & \prov{15.8} & \prov{48.9} & 4.7 \\
All & \prov{24{,}317} & \prov{19.8} & \prov{59.9} & \prov{45.1} & \prov{8.9} & \prov{26{,}638} & \prov{23.8} & \prov{65.7} & \prov{45.5} & \prov{7.9} & \prov{19{,}147} & \prov{43.4} & \prov{16.1} & \prov{48.3} & \prov{4.5} \\
\bottomrule
\end{tabular}
\end{table*}

\cref{supp:tab:models} lists the evaluated models, and \cref{supp:tab:gapfull} extends \cref{tab:gap} with SimLingo~\cite{simlingo}, with the normalized gap of \cref{supp:eq:norm} and with the baselines on the routes of Dev10.
SimLingo is trained on its own dataset and states its decision as a commentary, so it is not listed in \cref{tab:gap}.
Its say-exec gap of 21.0\% is the lowest of the three baselines, while its commentary and its own plan still disagree on 49.3\% of the frames for the speed decision and on 43.4\% for the yield decision.
The normalized gaps show that the small yield gaps of MindDrive and ORION do not indicate agreement.
Their normalized gaps are 0.84 and 0.97, so their stated decision and their planned trajectory agree barely more often than two independent outputs.
In open loop, on the 10{,}806 labeled validation frames, the say-do gap of \method is 0.4\% and its normalized gap 0.006.

\paragraph{Same routes.}
In \cref{tab:gap}, \method is evaluated on Dev10 and the baselines on all their logged routes.
The lower block of \cref{supp:tab:gapfull} evaluates all models on the ten routes of Dev10.
On these routes, the say-do gap of MindDrive is 80.1\% for the speed decision and 18.8\% for the yield decision, compared with 83.5\% and 19.8\% on all routes, and its say-exec gap is 80.2\%.
For ORION, the say-do and the say-exec gap of the yield decision are 29.4\% and 73.3\%.
On the same routes, the say-do gap of \method is 78.3 points below that of MindDrive, and its say-exec gap 27.8 points.
SimLingo again has the lowest say-exec gap, 19.4\%.

\paragraph{Which output is wrong.}
On the open-loop validation frames, the expert trajectory tells which of the two outputs is wrong.
The speed token of MindDrive matches the decision of the expert on \prov{33\%} of the frames, and its planned trajectory on 71\%.
For ORION, the yield decision of the language head matches the expert on \prov{49\%} of the evaluated frames, and its planned trajectory on \prov{80\%}.
For both models, the trajectory is the more accurate output, in line with \cref{sec:exp:main}.
In closed loop, the most frequent conflict of MindDrive is a token that maintains the speed while the plan accelerates, on 15{,}360 of its 24{,}317 logged frames.

\subsection{More Ablation Studies}
\label{supp:sec:ablation}\suppanchor{\SuppAblationRef}{\thesubsection}
\cref{tab:ablation} removes the \codebook, the counterfactual data or the consistency losses as a whole.
\cref{supp:tab:ablation} complements it in open loop with measures that \cref{tab:ablation} does not report: the collision rate, the distance between the trajectories of two decisions and the follow rate.
Every variant retrains the \codebook, the fusion MLP and, unless it is frozen, the speed branch on the same cached hidden states (\cref{supp:sec:training}), and the commitment is either the label of the frame or the output of a Decision Expert.
\begin{figure*}[t]
\centering
\includegraphics[width=\linewidth]{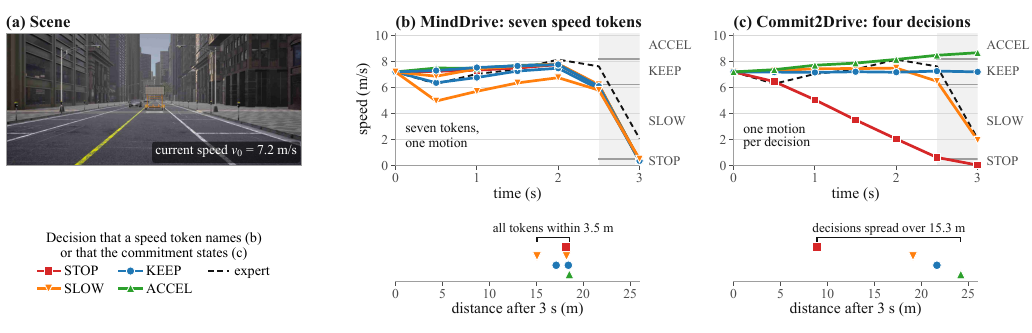}
\caption{\textbf{Candidate trajectories of MindDrive and of \method in the same scene of the validation set}, in which a construction site blocks the lane ahead. (a) Front camera image. (b) Speed over the next 3\,s of the seven candidate trajectories of MindDrive, one for each speed token, in the color of the decision that the token names. All seven brake in the last step and are read as \textsc{stop}, and they end within 3.5\,m of each other. The token that MindDrive states is \texttt{maintain\_moderate\_speed}. (c) Speed of the trajectories of \method under its four decisions. Each ends in the region of its own decision, and their end points lie up to 15.3\,m apart. Dashed: the expert. Gray band: the last 0.5\,s, whose mean speed is the end speed that $\Phi$ reads; the lines at 0.5\,m/s and at $v_0\pm1$\,m/s separate the four decisions. Bottom: distance traveled within the 3\,s; in (b), two of the seven markers coincide.}
\label{supp:fig:commitments}
\end{figure*}

\begin{table*}[t]
\begin{minipage}[t]{\columnwidth}
\centering
\caption{\textbf{Perception versus language for ORION in closed loop} (26{,}638 frames). Acc.: accuracy of the fact reported by the perception heads (see) and of the answer of the language head to a question about the same fact (say), against the privileged simulator state. See-say gap: share of frames on which the answer contradicts the perception of the model itself, on all, simple and complex frames. Say wrong: share of the scored disagreements in which the language head is wrong. All values in \%. The see-say gap of the hazard is scored on the 8{,}950 frames outside its tolerance band, and ``Red, given light'' on the 2{,}138 frames on which perception reports a light that governs the ego lane.}
\label{supp:tab:seesay}
\small
\setlength{\tabcolsep}{3.5pt}
\begin{tabular}{@{}lcccccc@{}}
\toprule
 & \multicolumn{2}{c}{Acc.} & \multicolumn{3}{c}{See-say gap} & Say \\
\cmidrule(lr){2-3}\cmidrule(lr){4-6}
Fact & see & say & all & simple & complex & wrong \\
\midrule
Hazard & \prov{95.5} & \prov{87.0} & \prov{12.9} & \prov{3.7} & \prov{14.8} & \prov{84} \\
Red light & \prov{95.7} & \prov{95.5} & \prov{7.1} & \prov{1.6} & \prov{7.4} & \prov{51} \\
Light governs ego & \prov{94.4} & \prov{94.2} & \prov{9.3} & \prov{3.5} & \prov{9.6} & \prov{51} \\
Red, given light & \prov{99.1} & \prov{98.8} & \prov{1.0} & \prov{0.0} & \prov{1.0} & \prov{62} \\
\bottomrule
\end{tabular}

\end{minipage}\hfill
\begin{minipage}[t]{\columnwidth}
\centering
\caption{\textbf{Rule-based functions that are plugged into \method without retraining.} Each rule reads the scene state $s$ from the perception of the model. When triggered, it issues the decision in place of the Decision Expert (\cref{eq:rule}).}
\label{supp:tab:rules}
\small
\setlength{\tabcolsep}{5pt}
\begin{tabular}{@{}l>{\raggedright\arraybackslash}p{3.2cm}l@{}}
\toprule
Rule & Trigger & Decision \\
\midrule
Traffic light & a red light governs the ego lane & \textsc{stop} or \textsc{slow} \\
Collision avoidance & a hazard is within the time-to-collision threshold & \textsc{stop} \\
\bottomrule
\end{tabular}

\end{minipage}
\end{table*}

\begin{figure*}[t]
\centering
\includegraphics[width=\linewidth]{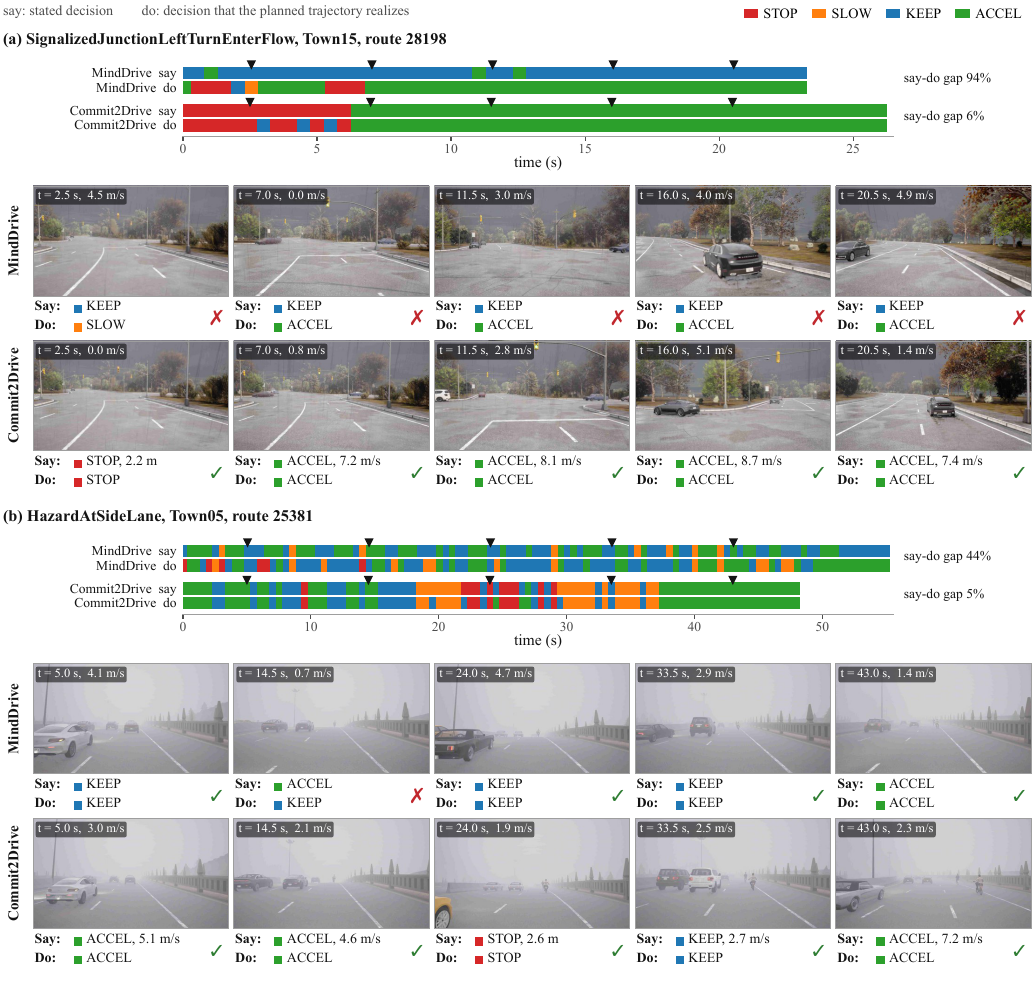}
\caption{\textbf{Closed-loop runs of MindDrive and \method on two routes of Dev10.} Ribbons: the stated decision (say) and the decision that the planned trajectory realizes (do) at 2\,Hz over the whole run, read with a horizon of 2.5\,s (\cref{supp:sec:protocol}). The triangles mark the five frames below, which are evenly spaced over the shorter of the two runs. Each frame is labeled with its time and the current speed. Under each frame: the two decisions, for \method with the target value of the commitment, and a check or a cross for their agreement. On both routes, each model completes the route with one collision (DS 60).}
\label{supp:fig:closedloop}
\end{figure*}

\begin{figure*}[t]
\centering
\includegraphics[width=\linewidth]{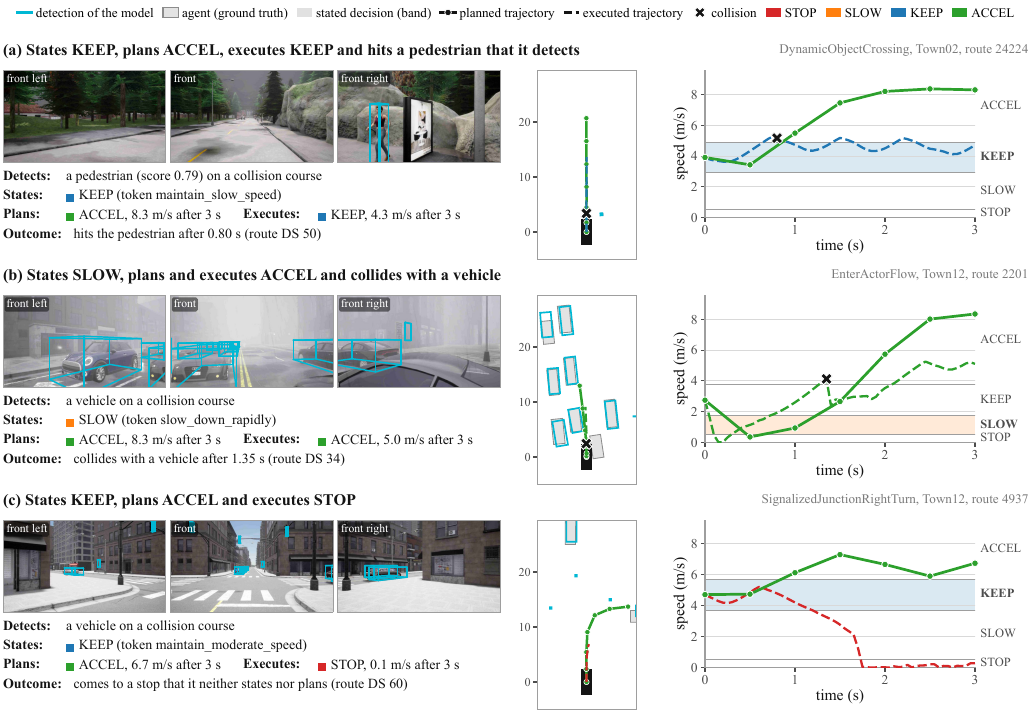}
\caption{\textbf{Cases in which the stated decision, the planned trajectory and the executed trajectory of MindDrive disagree}, from its closed-loop evaluation. Left: the three front cameras with the detections of the model. Middle: top view with the planned trajectory and the trajectory that is executed over the next 3\,s; forward is up, and the ticks are in meters. Right: speed over the next 3\,s; the band is the decision that the stated speed token names. The color of a band or of a curve shows its decision, which is read over these 3\,s and is the same with the horizon of 2.5\,s of \cref{supp:sec:protocol}. (a) The model detects a pedestrian, states a maintained speed and plans to accelerate, and hits the pedestrian after 0.80\,s. (b) It states a rapid slowdown, while its plan and the executed trajectory accelerate into a collision. (c) It states a maintained speed, plans to accelerate and comes to a stop.}
\label{supp:fig:cases}
\end{figure*}

\begin{figure*}[t]
\centering
\includegraphics[width=\linewidth]{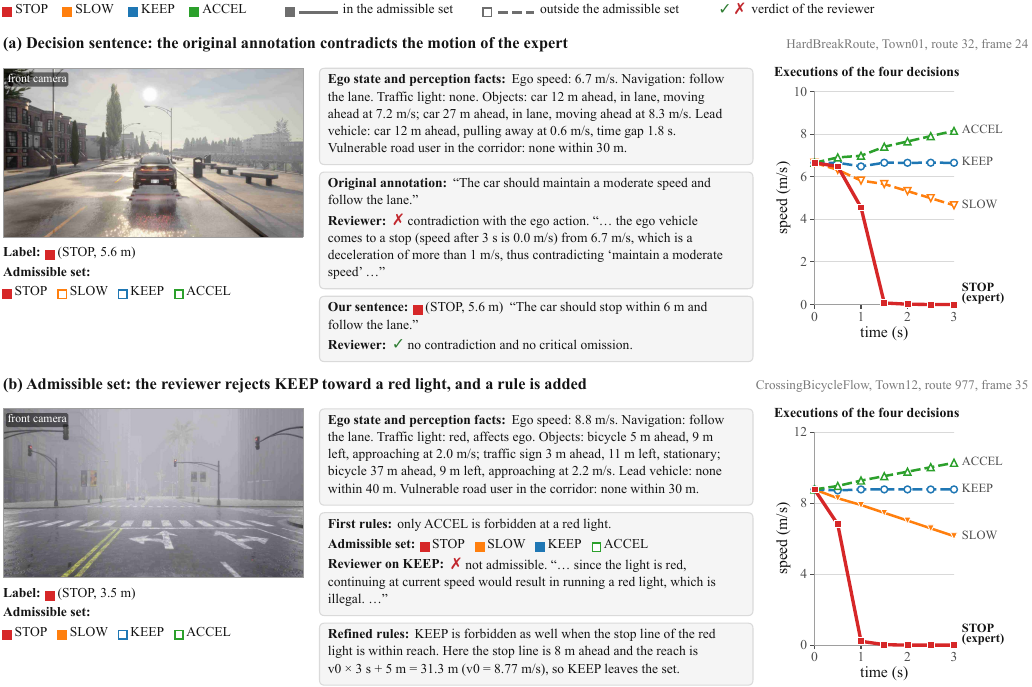}
\caption{\textbf{Records of the dataset before and after the review and refinement}, for two frames of the validation set. Left: the front camera image with the label and the admissible set of the final record. Middle: the text of the record with the verdict of the reviewer, of which an excerpt is quoted. Right: speed of the executions of the four decisions over the next 3\,s. (a) The original annotation states a maintained speed while the expert stops behind the vehicle ahead, which brakes hard in this scenario, and the reviewer finds the contradiction. Our sentence is rendered from the commitment that is read from the expert trajectory, and the reviewer has no objection. The executions of the other three decisions collide, so only \textsc{stop} is admissible. (b) Under the first version of the rules, \textsc{keep} is admissible although the stop line of a red light is 8\,m ahead, and the reviewer rejects it. The rule of \cref{supp:eq:reach} on the reach of a red light is added and removes \textsc{keep} from the admissible set, while \textsc{slow} remains admissible under the rules of \cref{supp:sec:pipeline}.}
\label{supp:fig:records}
\end{figure*}

The counterfactual data matters most.
With supervision on the decision of the expert only, the trajectories of two decisions end 2.65\,m apart instead of 8.39\,m, the trajectory follows every forced decision on 32.4\% of the frames instead of 99.0\%, and the collision rate under a label commitment rises from 0.50\% to 1.08\%.
Removing $\mathcal{L}_{\mathrm{ground}}$ and $\mathcal{L}_{\mathrm{sem}}$ leaves these measures almost unchanged, since $\mathcal{L}_{\mathrm{exec}}$ still supervises the generated trajectory and requires it to be read back as the decision.
The two losses supervise the code of the \codebook, which these measures do not capture.
Without them, the grounding head receives no supervision, and the nominal speed profile $\hat w$ that it decodes from the code no longer matches the decision.
On the training frames, it deviates from the speed profile of its target by 4.77\,m/s on average, compared with 0.22\,m/s with the two losses.
Freezing the speed branch raises the collision rate under a label commitment from 0.50\% to 0.60\%.

\subsection{Decision Interventions}
\label{supp:sec:intervention}
\paragraph{MindDrive and \method.}
\cref{supp:tab:follow} gives the follow rate of \cref{supp:eq:follow} for every decision.
A decision is evaluated on the frames on which an execution of it exists, between 2{,}509 and 3{,}429 of the 3{,}434 frames of the subset.
On the validation subset, the trajectory of MindDrive follows a forced speed token on 27\% to 30\% of the frames.
Its seven candidate trajectories fall into one decision on 86\% of the frames, and the candidates of two speed tokens that name different decisions end only 0.51\,m apart on average.
The token is an index without a kinematic meaning.
The trajectory of \method follows a forced decision on at least 99\% of the frames for each of the four decisions, and the trajectories of two decisions end 8.38\,m apart on average, close to the 8.49\,m of the executions that supervise them.
On the full validation set (10{,}806 labeled frames), the trajectory of MindDrive follows a forced speed token on 25\% to 36\% of the frames, and that of \method again on at least 99\%.

\paragraph{SimLingo.}
\cref{supp:tab:probes} lists the interventions on SimLingo~\cite{simlingo}.
Forcing the commentary to a decision moves the trajectory into this decision on \prov{32\%} of the frames on average, whereas an explicit instruction, the task that Action Dreaming trains, succeeds on \prov{94\%} to \prov{100\%}.
The trajectory therefore depends much less on the decision that the model itself states than on an external instruction.
The official agent of SimLingo keeps the dropout of its adapters active at inference, and we run the interventions with and without it.

\subsection{Consistency across Scene Types}
\label{supp:sec:strata}
Frames are assigned to scene types from the privileged simulator state and the scenario of the route.
Model outputs are not used, so the scene types are the same for all models.
A frame belongs to
\emph{C1} if the route turns by more than \prov{$15^\circ$} within the distance that the ego vehicle covers in 3\,s, at least 15\,m;
\emph{C2} if a pedestrian or cyclist is within \prov{20}\,m or inside the ego corridor;
\emph{C3} if a vehicle has a time to collision below \prov{3}\,s or cuts into the corridor;
\emph{C4} if the ego vehicle is slower than \prov{1}\,m/s, a red light or a stop sign of the ego lane is within \prov{30}\,m, or a stationary vehicle blocks the corridor within \prov{20}\,m;
\emph{C5} if a static obstacle is inside the corridor;
\emph{C6} at night or in rain;
and \emph{C7} if at least \prov{three} moving agents are within \prov{30}\,m and a vehicle is in the corridor.
A frame also belongs to C1, C2, C3, C5 or C7 if the scenario of its route is of this type.
A frame is \emph{simple} if none of the conditions on the simulator state holds and the ego vehicle drives at more than \prov{3}\,m/s, and \emph{complex} if it belongs to at least one scene type and is not simple.

\cref{supp:tab:strata} gives the gaps and the infraction rate of each scene type.
For MindDrive and ORION, the say-exec gap grows from 13.2\% and 9.1\% on simple frames to 63.3\% and 69.1\% on complex ones.
It peaks at 82.2\% and 87.8\% on stopped or low-speed frames (C4), the scene type with the highest infraction rate.
SimLingo shows the opposite order for the say-exec gap, with 28.1\% on simple frames and 15.8\% on complex ones, while its say-do gap more than doubles.

\subsection{Perception versus Language}
\label{supp:sec:seesay}

\method grounds the decision in facts that the perception of the model reports.
\cref{supp:tab:seesay} motivates this design with ORION~\cite{orion}, whose language head can be asked about the facts that its perception heads report.
We compare two facts in closed loop, a hazard and a red light, and list the two conditions of the red light separately.
A hazard is reported when a vehicle or a vulnerable road user, rolled out at constant velocity, meets the front of the ego vehicle within \prov{3}\,s.
A red light is reported when a traffic light governs the ego lane and is red.
Each disagreement is checked against the privileged simulator state with a tolerance band, and frames inside the band are not scored.
The answers of the language head contradict the perception of the model itself most often for the hazard, on 12.9\% of the frames on which this fact is scored and 14.8\% of the complex ones, and the language head is wrong in \prov{84\%} of the scored disagreements.
On the open-loop validation frames, the same analysis gives a see-say gap of \prov{0.4\%} on simple frames and \prov{14.9\%} on complex ones on average over the facts, with the language head wrong in \prov{94\%} of the scored disagreements.
Perception is the more reliable source for the hazard and as accurate as the language head for the red light, so \method builds the admissible set at inference from the perception facts.

\subsection{Rule-Based Decisions}
\label{supp:sec:rules}

\cref{supp:tab:rules} lists the two rule-based functions of \cref{sec:exp:ablation}.
A triggered rule issues the decision in place of the Decision Expert, and the Action Expert executes it without modification or retraining.
The experiment tests this extensibility, without evaluating the quality of the two rules.
On Dev10, the DS changes from 53.09 to 51.57 with the rules active.
In the Dev10 run that measures the latency of a rule-issued decision (\cref{supp:sec:runtime}), a rule issues the decision on 7{,}375 of the 26{,}209 steps (28.1\%), and the trajectory of the Action Expert realizes the rule-issued decision on 7{,}372 of them.
Given the hidden state $h_a$, the \codebook and the speed branch take a median of 14.7\,ms on one CPU thread for one commitment.

\subsection{More Qualitative Results}
\label{supp:sec:qualitative}

\paragraph{Decisions with distinct trajectories.}
\cref{supp:fig:commitments} compares the candidate trajectories of MindDrive and of \method in the same scene.
The seven candidates of MindDrive overlap, while every admissible decision of \method yields its own trajectory, which $\Phi$ reads back as that decision.

\paragraph{Closed-loop driving.}
On two routes of Dev10 (\cref{supp:fig:closedloop}), the decision that MindDrive states differs from the decision of its planned trajectory on 94\% and 44\% of the frames, compared with 6\% and 5\% for \method.

\paragraph{Failure cases of MindDrive.}
In the frames of \cref{supp:fig:cases}, the stated decision, the planned trajectory and the executed trajectory of MindDrive disagree.

\paragraph{Records of the dataset.}
\cref{supp:fig:records} shows the records of two validation frames before and after the review and refinement of \cref{supp:sec:pipeline}.
In the first frame, the original annotation states a maintained speed while the expert stops, and our sentence passes the review.
On 228 of the 600 reviewed frames, the reviewer finds such a contradiction in the original annotation and none in our sentence, and the reverse holds on 13 frames.
In the second frame, the reviewer rejects \textsc{keep} toward a red light, and the refined rules remove it.

\end{document}